\documentclass[runningheads]{llncs}

\usepackage[review,year=2024,ID=3472]{submit}
\usepackage{submitabbrv}

\usepackage{graphicx}

\usepackage{booktabs}
\usepackage{array}
\newcolumntype{P}[1]{>{\centering\arraybackslash}p{#1}}
\usepackage{tikz}
\usepackage{comment}
\usepackage{amsmath,amssymb} % define this before the line numbering.
\usepackage{bbm}
\usepackage{color}
\usepackage{colortbl}
\usepackage{multirow}
\usepackage{graphicx}
\usepackage{subcaption}

\usepackage{soul}

\newcommand{\xx}{\mathbf{x}}
\newcommand{\soft}{\mathbf{s}}
\newcommand{\sss}{\mathbf{p(y|x_i)}}
\newcommand{\yy}{\mathbf{y}}
\newcommand{\dd}{\mathbf{d}}

\newcommand{\llll}{\mathbf{l}}

\newcommand{\real}{\mathbb{R}}

\newcommand{\loss}{\mathcal{L}}
\definecolor{brightgray}{RGB}{220,220,220}

\newcommand{\appen}{\textcolor{red}{Appendix}\xspace}

\newcommand{\CC}[1]{\cellcolor{brightgray!#1}}

\usepackage{pifont}

\newrobustcmd*{\mytriangle}[1]{\tikz{\filldraw[draw=#1,fill=#1] (0,0) --
(0.1cm,0) -- (0.05cm,0.1cm);}}

\usepackage[accsupp]{axessibility}  % Improves PDF readability for those with disabilities.
\usepackage{tikz}

\usepackage[pagebackref,breaklinks,colorlinks,citecolor=submitblue]{hyperref}
\usepackage{orcidlink}

\begin{document}
\nolinenumbers
\title{Do not trust what you trust: Miscalibration in Semi-supervised Learning}
\maketitle

\begin{abstract}
State-of-the-art semi-supervised learning (SSL) approaches rely on highly confident predictions to serve as pseudo-labels that guide the training on unlabeled samples. An inherent drawback of this strategy stems from the quality of the uncertainty estimates, as pseudo-labels are filtered only based on their degree of uncertainty, regardless of the correctness of their predictions.  Thus, assessing and enhancing the uncertainty of network predictions is of paramount importance in the pseudo-labeling process. In this work, we empirically demonstrate that SSL methods based on pseudo-labels are significantly miscalibrated, and formally demonstrate the minimization of the min-entropy, a lower bound of the Shannon entropy, 
as a potential cause for miscalibration. To alleviate this issue, we integrate a simple penalty term, which enforces the logit distances of the predictions on unlabeled samples to remain low, preventing the network predictions to become overconfident. Comprehensive experiments on a variety of SSL image classification benchmarks demonstrate that the proposed solution systematically improves the calibration performance of relevant SSL models, while also enhancing their discriminative power, being an appealing addition to tackle SSL tasks. The code is publicly available at \url{https://github.com/ShambhaviCodes/miscalibration-ssl}.
\end{abstract}

\section{Introduction}
\label{sec:intro}
Deep learning models have significantly advanced the state-of-the-art across a myriad of tasks \cite{masana2022class,minaee2021image}. %, including image classification \cite{masana2022class}, or segmentation \cite{minaee2021image}, for example. %partially due to the access of large scale datasets. 
Nonetheless, their success has been often contingent on the availability of large amounts of labeled data. Having access to curated large training datasets, however, is not easy, and often involves a tremendous human labor, particularly in those domains where labeling data samples requires expertise, hindering the progress to address a broader span of real-world problems. %, such as healthcare. %, for example. 

% pseudo-labeling that aims to effectively expand the labeled data by generating predicted pseudo-labels for the unlabeled samples and then using them as ground-truth labels for model training.

% The success of pseudo-label based SSL algorithms hinges on the quality of the pseudo-labels...a more straightforward solution is using pseudo-labels to periodically learn from the model itself to encourage entropy minimization... 

% Numerous SSL algorithms have been introduced, with one of the most prevalent assumptions being entropy minimization, which requires the decision boundaries to lie in low density areas Wang et al. (2022a). In order to achieve this, pseudo-labels are introduced in the context of SSL
Semi-supervised learning (SSL) \cite{chapelle2006ssl} mitigates the need for large labeled datasets by providing means of efficiently leveraging unlabeled samples, which are easier to obtain. This %appealing 
learning paradigm has led to a plethora of approaches, which can be mainly %roughly 
categorized into consistency regularization \cite{bachman2014learning,laine2016temporal} and pseudo-labeling \cite{lee2013pseudo,xie2020self} methods. Indeed, state-of-the-art SSL approaches \cite{wang2023freematch,zhang2021flexmatch,sohn2020fixmatch,chen2023softmatch,zheng2022simmatch} combine both strategies, obtaining promising results. The underlying idea of these approaches follows the low-density and smoothness assumptions in SSL \cite{chapelle2005semi}. In particular, incorporating pseudo-labels from unlabeled data points into the training process aids the decision boundary to lie in low density regions. Furthermore, consistency regularization assumes that the same unlabeled data point should yield the same pseudo-label regardless of the perturbations applied, implicitly capturing the underlying data manifold. %encourages the model to make predictions that are consistent with this underlying manifold structure. 
As the model can produce very uncertain predictions for strongly perturbed samples, these techniques incorporate a threshold, either fixed \cite{sohn2020fixmatch} or adaptive \cite{wang2023freematch,zhang2021flexmatch}, to only integrate very confident samples in the training loss. Thus, all samples whose predicted probabilities are highly confident are trusted by these methods as supervisory signals for subsequent steps, even when their predictions are wrong. % ....only retain an artificial label if the model assigns a high probability to one of the possible classes
 
Despite being a standard practice, recent evidence %in the literature 
\cite{chen2023softmatch} suggests that the amount of incorrect pseudo-labels integrated into the training is not negligible, potentially undermining the optimization process. Hence, given that the generated pseudo-labels play a significant role in the training of SSL models, producing accurate uncertainty estimates is of pivotal importance. %\jose{maybe talking here about the problem of minimizing entropy...}
Nevertheless, while we have observed a remarkable progress in their discriminative performance, little attention has been paid to studying, and improving, the calibration of SSL approaches. Motivated by these findings, in this work we address the critical yet under-explored issue of miscalibration in SSL, particularly for those methods based on pseudo-labeling. To this end, we select a set of relevant and recent strategies that build on %based on the combination of 
pseudo-labels and consistency regularization \cite{sohn2020fixmatch,wang2023freematch,zhang2021flexmatch} and empirically demonstrate that they are poorly calibrated. Furthermore, we explore the underlying causes of this issue and shed light about the potential reasons that produce overconfident pseudo-labeling SSL models. Last, inspired by these observations we propose a simple solution to tackle miscalibration in these models. Our contributions can be therefore summarized as follows: 

\begin{enumerate}
    \item We empirically demonstrate that state-of-the-art SSL approaches based on pseudo-labels are significantly miscalibrated. Through our analysis, we formally show that the cause of miscalibration is the minimization of a min-entropy term, a specific case of the Rényi family of entropies, on a considerably large proportion of unlabeled samples, which forces the model to yield overconfident predictions. Indeed, the ensuing gradients from this term strongly push the unlabeled samples to be highly confident since the beginning of the training, even if their class predictions are incorrect. This results in large logit magnitudes, a phenomenon known to cause miscalibration.  

    \item Based on our observations, we propose to use a simple solution that refrains the model from pushing unlabeled samples towards very unconfident regions, improving the calibration %performance 
    of pseudo-labeling 
    SSL methods. %based on pseudo-labels. %, as well as their discriminative performance. 
    More concretely, we add a penalty term on a dominant set of unlabeled samples, which enforces logit distances to remain low, alleviating the miscalibration issue. 

    \item Through a comprehensive set of experiments, we empirically demonstrate that the proposed approach consistently improves the uncertainty estimates of a set of very relevant and recent state-of-the-art pseudo-labeling SSL approaches on popular benchmarks, in both standard and long-tailed classification tasks. In addition, in most cases, the proposed solution further improves their discriminative performance. 
\end{enumerate}

\section{Related work}
\subsection{Semi-supervised learning}

%Prevailing semi-supervised learning (SSL) strategies include pseudo-labeling and consistency regularization. 

Prevailing semi-supervised learning (SSL) approaches heavily rely on the concept of pseudo-labels \cite{lee2013pseudo,shi2018transductive} and consistency regularization \cite{bachman2014learning,laine2016temporal,sajjadi2016regularization,tarvainen2017mean,miyato2018virtual}, where labels are dynamically generated for unlabeled data throughout the training process. Essentially, these methods exploit the role of perturbations, by stochastically perturbing the unlabeled images and enforcing consistency across their predictions. This consistency is achieved by a pseudo-supervised loss, where the predictions over the strong perturbations are supervised by the pseudo-labels obtained from the weak perturbations \cite{xie2020unsupervised,wang2023freematch,zhang2021flexmatch,chen2023softmatch,zheng2022simmatch,yang2023shrinking,xu2021dash,zheng2022simmatch}. This paradigm to use artificial labels facilitates the integration of unlabeled data into the learning process, thereby augmenting the training set and improving the model ability to generalize. To avoid introducing noise in the pseudo-supervision process, these approaches retain a given pseudo-label only if the model assigns a high probability to one of the possible classes. This strategy effectively harnesses the information from unlabeled data by leveraging the network confidence in assigning pseudo labels, enabling the model to access the valuable knowledge encapsulated within these unlabeled samples. Thus, the main differences across the different approaches based on pseudo-labels lie on the mechanism introduced to select confident samples. For example, FixMatch \cite{sohn2020fixmatch} and ShrinkMatch \cite{yang2023shrinking} employ a fixed threshold, % $\tau$, 
whereas Dash \cite{xu2021dash} proposes a dynamically growing threshold. Other approaches, such as FlexMatch \cite{zhang2021flexmatch} and FreeMatch \cite{wang2023freematch}, integrate class-adaptative thresholds, considering a larger amount of unlabeled data which is otherwise ignored, especially at the early stage of the training process. %(by leveraging the model predictions during training)
%\textcolor{red}{More recently, SoftMatch \cite{chen2023softmatch} presented a truncated Gaussian function that weights the samples based on their confidence, which can be viewed as a soft version of the confidence threshold, as it not only considers highly confident samples, but also data points whose predictions are more uncertain.} 

%Check this: UDA, MixMatch and ReMixMatch (do not do pseudo-labels but sharpening: sharpen the artificial label to encourage the model to produce high-confidence predictions).

\noindent \textit{Limitations of pseudo-labeling SSL from a calibration standpoint.} Although these methods enhance the discriminative power of deep models, their calibration %performance 
has been significantly overlooked, lacking of principled strategies to simultaneously improve the classification performance while maintaining the quality of the uncertainty estimates. As pseudo-labeling state-of-the-art SSL approaches trust highly confident artificial labels derived from unlabeled samples, understanding how this confidence is assigned, and ensure its accuracy, %and whether it is correctly assigned, 
is of paramount importance. Very recently, only BAM \cite{loh2023mitigating} studied miscalibration in SSL, and proposed to replace the last layer of a neural network by a Bayesian layer. Nevertheless, the source of miscalibration was not explored, and in-depth empirical results were not reported.%in SSL.

%Differentiate between entropy regularization (pseudo-labels) and consistency-based. 

%In entropy-based: hard vs soft pseudo-labels. FixMatch \cite{sohn2020fixmatch},  UDA \cite{xie2020unsupervised}, FlexMatch \cite{zhang2021flexmatch}, FreeMatch \cite{wang2023freematch}, SoftMatch \cite{chen2023softmatch}

\subsection{Calibration}
% Highlight that all of them focus on supervised approaches...
Recent evidence \cite{guo2017calibration,muller2019does,mukhoti2020calibrating} has shown that
deep networks are prone to make overconfident predictions due to miscalibrated output probabilities. This emerges as a byproduct of minimizing the prevalent cross-entropy loss, which occurs when the softmax predictions for all training samples fully match the ground-truth labels, and thus the entropy of output probabilities is encouraged to be zero. To mitigate the miscalibration issue, and to better estimate the predictive uncertainty of deterministic models, two main families of approaches have emerged recently: \textit{post-processing} \cite{guo2017calibration,ding2021local,tomani2021post} and \textit{learning} \cite{pereyra2017regularizing,muller2019does,mukhoti2020calibrating,liu2022devil,cheng2022calibrating,liu2023class,noh2023rankmixup,murugesan2023trust,larrazabal2023maximum,park2023acls} approaches. Among the post-processing strategies, Temperature Scaling (TS) \cite{guo2017calibration} has been a popular alternative, which manipulates logit outputs monotonely, by applying a single scalar temperature parameter. This idea is further extended in \cite{ding2021local}, where a local TS per pixel is provided by a regression neural network. Nevertheless, despite the simplicity of these methods, learning approaches have arisen as a more powerful choice, as in this scenario the model adapts to calibration requirements alongside its primary learning objectives, optimizing both aspects simultaneously. Initial attempts integrated learning objectives that maximize the entropy of the network softmax predictions either explicitly \cite{pereyra2017regularizing,larrazabal2023maximum}, or implicitly \cite{mukhoti2020calibrating,muller2019does,cheng2022calibrating}. In order to alleviate the non-informative nature of simply maximizing the entropy of the softmax predictions, recent work \cite{liu2022devil,liu2023class} has presented a generalized inequality constraint, which penalizes logits distances larger than a pre-defined margin. %More recently, \cite{park2023acls} presented ACLS, an adaptive and conditional Label Smoothing approach, which uses an adaptive smoothing function (non-uniformly according to logit values)

\section{Semi-supervised learning and calibration}

\subsection{Problem statement}

In the semi-supervised learning scenario, the training dataset is composed of labeled and unlabeled data points. In this setting, let $\mathcal{D}_L = \{(\xx_i,\yy_i)\}_{i}^{N_L}$ be the labeled dataset and $\mathcal{D}_U = \{\xx_i\}_{i}^{N_U}$ the unlabeled dataset, where $N_L$ and $N_U$ represent the number of labeled and unlabeled samples, respectively, and $N_L<<N_U$. Furthermore, 
$\xx_i \in \real^d$ is a $d$-dimensional training sample, with $\yy_i~\in~\{0,1\}^K$ its associated ground truth (only for labeled data points, i.e., $\xx_i \in \mathcal{D}_L$) that assigns one of the $K$ classes to the sample. The objective is, given a batch of labeled and unlabeled samples, to find an optimal set of parameters of a deterministic function, e.g., a neural network, parameterized by $\boldsymbol{\theta}$, by using a compounded loss including a labeled and an unlabeled term. Given an input image $\xx_i$, the network will generate a vector of logits  $f_{\boldsymbol{\theta}}(\xx_i)=\llll_i \in \real^K$, which can be converted to probabilities %prediction vector $\pp_i$ 
with the softmax function. %, $\sss_i=\sigma(\llll_i)$.

\noindent \textbf{Supervised loss.} The supervised objective is typically formulated as a standard cross-entropy $\mathcal{H}$ between the one-hot encoded labels $\yy_i$ and the corresponding softmax predictions $\mathbf{p(y|x_i)}\in[0,1]^K$ of labeled samples:

\begin{align}
\label{eq:ce}
    \loss_{S} = \mathcal{H}(\yy_i,\sss) = - \sum_{i \in \mathcal{D}_L} \yy_i \log \sss
\end{align}

\noindent \textbf{Unsupervised loss.} Most modern SSL approaches adopt a consistency regularization strategy based on pseudo-labeling for the unsupervised objective. %, which follows the continuity assumption of SSL \cite{}. 
%More concretely, these methods resort to a confidence-based thresholding mechanism to mask out the unconfident, and possibly incorrect, pseudo-labels from training. 
To this end, the same image $\xx_i$ follows a set of weak and strong augmentations, denoted as $\omega (\cdot)$ and $\Omega (\cdot)$, respectively. Thus, a pseudo-cross-entropy term on the unlabeled training dataset can be formulated as:

%%%% JOSE - HERE %%%%
\begin{align}
\label{eq:pseudo-ce}
    \loss_{U} = - \sum_{i \in \mathcal{D}_U} \tilde{\yy}_i \log \mathbf{p}(\yy|\Omega(\xx_i))
\end{align}

\noindent where $\tilde{\yy}_i$ is the one-hot encoding of the $\arg \max$ of the softmax probabilities for the weak augmented version, i.e., $\arg \max (\mathbf{p}(\yy|\omega(\xx_i))$. To avoid that samples with high uncertainty, and possibly incorrect predictions, intervene in the optimization of the term in \cref{eq:pseudo-ce}, a common strategy is to retain only discrete pseudo-labels whose largest class probability fall above a predefined threshold \cite{lee2013pseudo,wang2023freematch,sohn2020fixmatch,zhang2021flexmatch}. %More concretely, these methods resort to a confidence-based thresholding mechanism to mask out the unconfident, and possibly incorrect, pseudo-labels from training.
This results in the following objective:

\begin{align}
\label{eq:weighted-pseudo-ce}
    \loss_{U} =  - \sum_{i \in \mathcal{D}_U} \mathbbm{1}(\max(\mathbf{p}(\yy|\omega(\xx_i))) \geq \tau)\tilde{\yy}_i \log \mathbf{p}(\yy|\Omega(\xx_i))
\end{align}

\noindent where $\tau$ is the predefined threshold. Note that all loss terms are normalized by the cardinality of each set, which we omit for simplicity.

%Detail how all these models are trained: a common formulation for all of them, highlighting what they do for weak and strong augmentations.
%Training SSL models

\subsection{Revisiting the calibration of semi-supervised models}
\label{ssec:cal-SSL}

We now introduce a series of observations revealing several intrinsic properties of semi-supervised methods built up on pseudo-labels generation, which allows us to motivate a calibration technique tailored for pseudo-label based SSL.

\begin{figure*}[ht!]
         \centering
         \includegraphics[width=\linewidth]{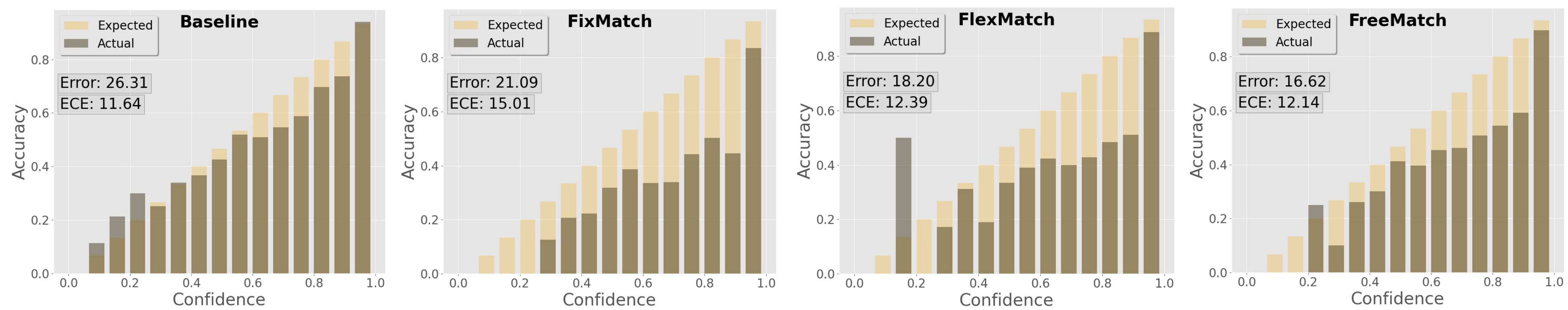}
     \caption{\textbf{Observation 1.} Reliability plots for a baseline supervised model (trained with \cref{eq:ce}) and three representative SSL approaches (trained with \cref{eq:start-eq}) on CIFAR-100. %(\textit{top}) and STL10 (\textit{bottom}). 
     These plots empirically highlight the calibration degradation observed when training with the standard unsupervised loss, %in \cref{eq:weighted-pseudo-ce}, 
     despite the gains achieved in discrimination. }%discriminative performance.} %Show this for baseline (i.e., trained only on labeled data) and the four SSL models. Show also a plot with the evolution of the ratio between number of samples with same and different argmax labels. You can pick just one dataset and one configuration (we will show all in supplemental material)}
     \label{fig:obs1}
\end{figure*}

\noindent \textbf{Observation 1. Semi-supervised learning degrades the calibration performance.} An important body of literature on SSL %semi-supervised learning 
relies on pseudo-labeling to leverage the large amount of unlabeled samples. To achieve this, a very common strategy is to generate weak augmentations of each unlabeled image, whose predictions serve as supervision for their strong augmentation counterpart, as presented in \cref{eq:weighted-pseudo-ce}. Pseudo-labeling \cite{lee2013pseudo} is indeed closely related to entropy regularization \cite{grandvalet2004semi}, which favors a low-density separation between classes, a commonly assumed prior for semi-supervised learning. While minimizing the entropy of the predictions can actually improve the discriminative performance of neural networks, it inherently favours overconfident predictions, which is one of the main causes of miscalibration. Figure \ref{fig:obs1} brings empirical evidence about this observation, where we can observe that \textit{despite bringing performance gains, in terms of accuracy, pseudo-label based SSL methods naturally degrade the calibration properties of a supervised baseline} trained with a few labeled samples.

%\begin{figure*}[ht!]
%\begin{center}   \includegraphics[width=0.99\linewidth]{images/observations.png}
%\end{center}
%   \caption{\textbf{Observation 1.} Reliability plots for a baseline supervised model (trained with \cref{eq:ce}) and three representative semi-supervised approaches (trained with \cref{eq:weighted-pseudo-ce}) on XXX dataset. These plots empirically confirm the calibration degradation when training with the standard unsupervised loss in \cref{eq:weighted-pseudo-ce}.} %Show this for baseline (i.e., trained only on labeled data) and the four SSL models. Show also a plot with the evolution of the ratio between number of samples with same and different argmax labels. You can pick just one dataset and one configuration (we will show all in supplemental material)}
%\label{fig:entropy}
%\end{figure*}

\noindent \textbf{Observation 2. Pseudo-labeling in SSL indeed minimizes min-entropy on unlabeled points.} The use of a hard label makes pseudo-labeling closely related to entropy minimization \cite{grandvalet2004semi}. Nevertheless, the different transformations that unlabeled images follow in modern SSL methods produce different probability distributions for the weak and strong versions of the same image, where the predictions of the former are used to correct the predictions of the later. Note that this is slightly different from the traditional pseudo-labeling approaches, 
where the same image is used for assigning the pseudo-label and updating the predictive model. %where images do not follow any transformation, and hence only one soft prediction per image is generated. 
Thus, albeit they are related, the minimization of entropy cannot be attributed as the cause of miscalibration. Motivated by this, we explore in this section the implications of the standard SSL unlabeled loss based on pseudo-labels and its effect on network calibration. The common learning objective is composed of two terms: the first one is the standard cross-entropy (CE) on labeled samples (\cref{eq:ce}), while the second term is a CE between pseudo-labels obtained from weak augmentations and the predictions of their strong augmented counterparts (\cref{eq:weighted-pseudo-ce}): 

\begin{align}
\label{eq:start-eq}
    \loss_T = \underbrace{- \sum_{i \in \mathcal{D}_L} \yy_i \log \soft_i}_{\textrm{CE on labeled samples}} \underbrace{- \sum_{i \in \mathcal{D}_{U}} \tilde{\yy}^w_i \log \soft^s_i}_{\textrm{Pseudo-CE on $\mathcal{D}_{U}$}}, 
\end{align}

%\noindent with $\soft_i=\sss$ being used for simplicity, and the superscripts $w$ and $s$ denoting weak and strong transformations, respectively.
\noindent where $\soft_i=\sss$ is used for simplicity, and the superscripts $w$ and $s$ denote weak and strong transformations, respectively. %From our observations, we found that, for a considerable large amount of unlabeled samples used in these losses, the category used as pseudo-label is the same as the one predicted on the strong augmented image. 
We now split the unlabeled dataset into $\mathcal{D_{U'}}$, which contains the unlabeled samples whose predicted class from weak and strong augmentations are different, i.e., $\arg \max (\soft^w_i) \neq \arg \max (\soft^s_i)$ and $\mathcal{D_{U''}}$, containing the samples whose predicted class from weak and strong augmentations are the same. %, i.e., $\arg \max (\soft^w_i) = \arg \max (\soft^s_i)$. 
Thus, we can decompose the right-hand term in \cref{eq:start-eq} into two terms, one acting over $\mathcal{D_{U'}}$ and one over $\mathcal{D_{U''}}$:

\begin{align}
\label{eq:tot1}
    \loss_T = \underbrace{- \sum_{i \in \mathcal{D}_L} \yy_i \log \soft_i}_{\textrm{CE on labeled samples}} \underbrace{- \sum_{i \in \mathcal{D}_{U'}} \tilde{\yy}^w_i \log \soft^s_i}_{\textrm{Pseudo-CE on $\mathcal{D}_{U'}$}} - \sum_{i \in \mathcal{D}_{U''}} \tilde{\yy}^w_i \log \soft^s_i%\underbrace{- \sum_{i \in \mathcal{D}_{U''}} \tilde{\yy}^w_i \log \soft^s_i}_{\textrm{Pseudo-label on $\mathcal{D}_{U''}$}}
\end{align}

In the above equation, the second term can be seen as a pseudo cross-entropy on $\mathcal{D}_{U'}$. Furthermore, as $\tilde{\yy}^w_i$ is equal to the one-hot vector from $\arg \max(\soft^s_i)$ on samples from $\mathcal{D}_{U''}$, the last term is equivalent to the min-entropy\footnote{Pseudo-label $\hat{y}_{i,c}^w=1$ if $\text{s}^s_{i,c}=\max_k \text{s}^s_{i,k}$, and 0 otherwise.}: %, which is a lower bound of the Shannon Entropy (See Figure \ref{fig:entropy-sub}, \textit{left}). Maybe explain here a bit about Renyi entropies...

\begin{align}
\label{eq:tot2}
    \loss_T = \underbrace{- \sum_{i \in \mathcal{D}_L} \yy_i \log \soft_i}_{\textrm{CE on labeled samples}} \underbrace{- \sum_{i \in \mathcal{D}_{U'}} \tilde{\yy}^w_i \log \soft^s_i}_{\textrm{Pseudo-CE on $\mathcal{D}_{U'}$}} \underbrace{- \sum_{i \in \mathcal{D}_{U''}} \log (\max_k \text{s}^s_{i,k})}_{\textrm{min-entropy on $\mathcal{D}_{U''}$}}
\end{align}

\begin{figure}[ht!]
    \centering
    \includegraphics[width=0.98\linewidth]{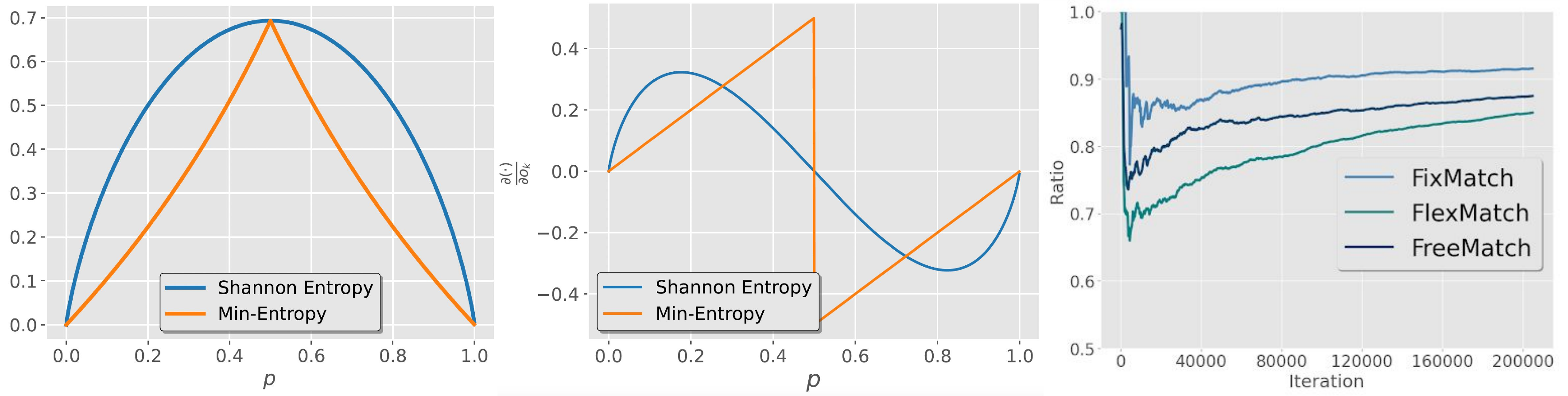}
     \caption{\textbf{Observation 2.} %(\textit{Left}) 
     \textbf{(Left)}{The unsupervised term in pseudo-label SSL is (approximately) equivalent to min-entropy, %which is 
     a lower bound of the Shannon Entropy. \textbf{(Middle)} Compared to the Shannon Entropy, the min-entropy is more aggressive in the gradient dynamics, particularly at the beginning of the training, when most predictions are %still 
     uncertain. \textbf{(Right)} Ratio of samples with same hard prediction for weak and strong augmentations %$\arg \max (\soft^w_i) = \arg \max (\soft^s_i)$ 
     that were above the selection threshold of three relevant SSL methods.}}
     \label{fig:entropy-sub}
\end{figure}

As shown in Figure \ref{fig:entropy-sub} in the case of a two-class distribution $(p,1-p)$, min-entropy is a lower bound of the Shannon Entropy, which has several implications in network miscalibration. In particular, while both Shannon Entropy and min-entropy reach their minimum at the vertices of the simplex, i.e., when $p~=~0$ or $p~=~1$ (\textit{left}), the dynamics of the gradients are different (\textit{middle}). More concretely, in the case of the Shannon entropy, the gradients of low-confidence predictions at the middle of the simplex are small and, therefore, dominated by the other terms at the beginning of training. %This can be, for example, the case of the supervised term in \cref{eq:} 
In contrast, by minimizing the min-entropy, the inaccuracies resulting from uncertain predictions are reinforced, i.e., pushed towards the simplex vertices, yielding early errors in the predictions, which are hardly recoverable, and potentially misleading the training process.

%\begin{figure*}[ht!]
%     \begin{subfigure}[b]{0.32\linewidth}
     
%         \centering
         %\includegraphics[width=\linewidth]{images/Entropies.pdf}
         % \caption{}
         % \label{fig:sum-rank}
   %  \end{subfigure}
   %  \begin{subfigure}[b]{0.32\linewidth}
     
  %       \centering
         %\includegraphics[width=\linewidth]{images/Derivatives.pdf}
         % \caption{}
         % \label{fig:sum-rank}
    % \end{subfigure}
   %  \begin{subfigure}[b]{0.32\linewidth}
   %      \centering
         %\includegraphics[width=\linewidth]{images/ratio.pdf}
         % \caption{}
         % \label{fig:sum-rank}
   %  \end{subfigure}
  %   \caption{\textbf{Observation 2.} %(\textit{Left}) 
  %   \textbf{(Left)} \textcolor{blue}{The unsupervised term in pseudo-label SSL is (approximately) equivalent to min-entropy, %which is 
  %   a lower bound of the Shannon Entropy. \textbf{(Middle)} Compared to the Shannon Entropy, the min-entropy is more aggressive in the gradient dynamics, particularly at the beginning of the training, when most predictions are %still 
  %   uncertain. \textbf{(Right)} Ratio of samples with same hard prediction for weak and strong augmentations %$\arg \max (\soft^w_i) = \arg \max (\soft^s_i)$ 
  %   that were above the selection threshold of three relevant SSL methods.}}
 %    \label{fig:entropy-sub}
%\end{figure*}

Furthermore, we empirically observed that the amount of samples where the \textit{$\arg \max$} of the predictions from weak and strongly augmented versions was the same, i.e., $\mathcal{D}_{U''}$, was significantly larger than those with different predictions, i.e., $\mathcal{D}_{U'}$ (Figure \ref{fig:entropy-sub}, \textit{right}). %found that, for a considerable large amount of unlabeled samples used in these SSL losses, the category used as pseudo-label is the same as the one predicted on the strong augmented image.... However, we empirically observed that the amount of samples where the \textit{$\arg \max$} of the predictions from weak and strongly augmented versions was the same, was significantly larger than those with different labels. Figure
In particular, as shown in this figure, and after some iterations, \textbf{more than 80\% of unlabeled samples included in the training share the same pseudo-label between weak and strong annotations,} regardless of the approach analyzed. Thus, based on our observations we argue that \textit{the training of SSL methods based on pseudo-labels can be approximated by a supervised term coupled with a regularization loss that minimizes the min-entropy of unlabeled samples.} This means that, while implicitly, or explicitly, minimizing the Shannon entropy on the network softmax predictions is known to cause miscalibration \cite{guo2017calibration}, employing the min-entropy aggravates the problem, which explains why SSL methods based on pseudo-labels are not well calibrated, particularly compared to a simple supervised baseline (Figure \ref{fig:obs1}).

\begin{figure*}[ht!]
\centering
     \begin{subfigure}[b]{0.4\linewidth}
         \centering
         \includegraphics[width=\linewidth]{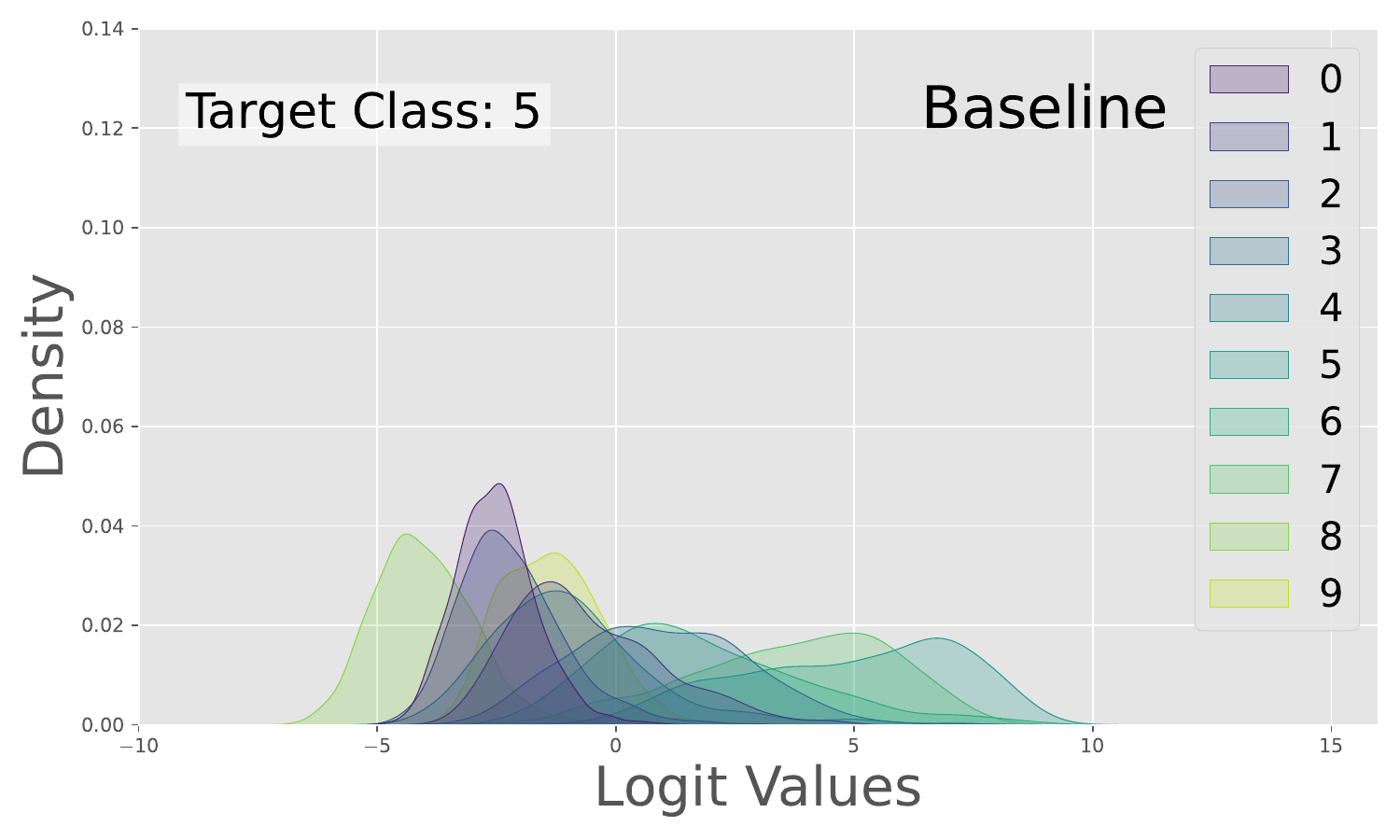}
         %\caption{Baseline}
         % \label{fig:sum-rank}
     \end{subfigure}
     \begin{subfigure}[b]{0.4\linewidth}
         \centering
         \includegraphics[width=\linewidth]{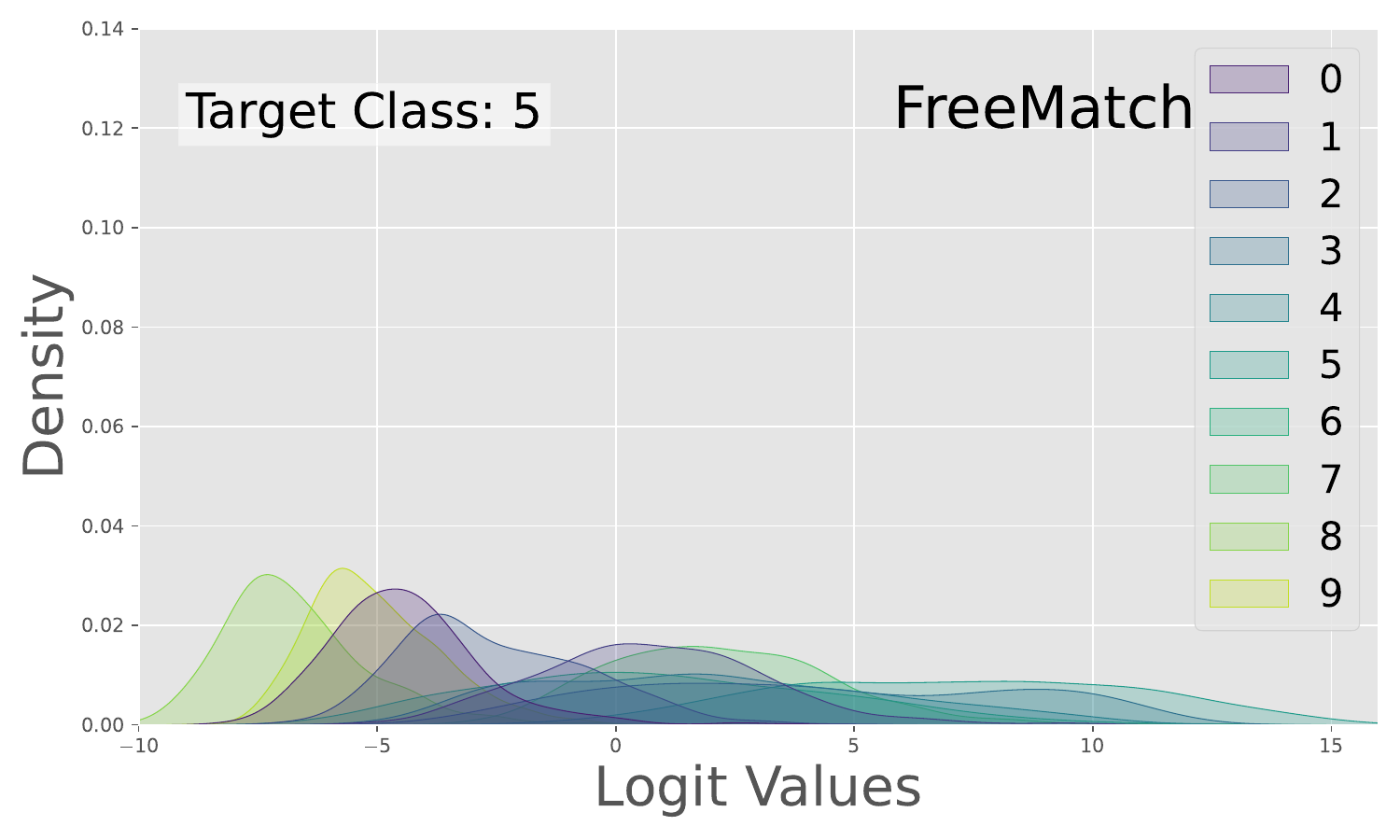}
         %\caption{FreeMatch}
         % \label{fig:sum-rank}
     \end{subfigure}
     \caption{\textbf{Observation 3.} These plots depict the Kernel Density Estimation of the logit distributions obtained by \textit{(left)} the supervised baseline trained with \cref{eq:ce} and \textit{(right)} FreeMatch on STL-10, % (40 labeled images), 
     for the samples belonging to class 5. We can observe that, even for non-target classes (k$\neq$5), the logit magnitudes in FreeMatch are larger, which translates to higher overconfidence in both correct and incorrect predictions. We select STL-10 due to its number of classes (10 \textit{vs.} 100 in CIFAR-100).}
     %\vspace{-10mm}
     \label{fig:obs3}
\end{figure*}

\noindent \textbf{Observation 3. Pseudo-label SSL techniques produce highly overlapped logit distributions, with large logit magnitudes and distances.} A direct implication of \textbf{Observation 2} is that softmax predictions in pseudo-labeling SSL methods become highly confident, which translates into larger logit values compared to a supervised baseline. As a result, even when a predicted category is incorrect, the network will still express a high level of certainty in its prediction. Take for instance the example shown in Figure \ref{fig:obs3}, which depicts the logit value distributions for the samples belonging to class 5. We can observe that, in the case of the supervised baseline, the maximum logit values for incorrect predicted classes is at $\approx$ 7.5 (classes 6 and 7) and around 6.5 (class 8). In contrast, when adding unlabeled samples in the form of pseudo-labels (e.g., in FixMatch \cite{sohn2020fixmatch}), the maximum logit values for incorrect classes increase to nearly 12.5. This means that, while both models yield incorrect predictions, these are highly confident in the SSL methods. Furthermore, the logit range in the supervised baseline goes from around -7 to 10, whereas FreeMatch produces logits in the range from -10 to 15, which will result in highest probability scores for the predicted category. From these observations, we argue that \textit{a well-calibrated SSL model should decrease the magnitude of logits associated with incorrect classes, as well as their total logits range, thereby decreasing confidence in erroneous predictions, while simultaneously preserving high values in logits corresponding to the target class.}

%\begin{figure}[ht!]
%\begin{center}   \includegraphics[width=0.90\linewidth]{images/Entropies.pdf}
%\end{center}
%   \caption{Entropies. The unsupervised term in SSL is (approximately) equivalent to min-entropy, which is a lower bound of the Shannon Entropy (more aggressive in the gradient dynamics). We can also show the gradients for both...}
%\label{fig:entropy}
%\end{figure}

%\textit{i)} better separate the target and non-target class logits and \textit{ii)} decrease the magnitude of the logits, to avoid over-confident predictions.  

%\begin{figure}[ht!]
%  \centering
%  \begin{minipage}[b]{0.4\textwidth}
%    \includegraphics[width=\linewidth]{images/Entropy.png}
%    \caption{Visual motivation (what minimizing entropy (or min-entropy) does.) \put(0,0){\circle{10.6}}\put(0,0){\color{red}\circle*{10}}}
%  \end{minipage}
%  \hfill
%  \begin{minipage}[b]{0.4\textwidth}
%    \includegraphics[width=\linewidth]{images/Expected.png}
%    \caption{What we want.}
%  \end{minipage}
%\end{figure}

\vspace{-24pt}

\section{Our solution}

%Examples of other simple approaches: Fan Y, Kukleva A, Dai D, Schiele B. Revisiting consistency regularization for semi-supervised learning. International Journal of Computer Vision. 2023 Mar;131(3):626-43.

Based on our findings, especially in \textbf{Observation 3}, it becomes evident that an effective strategy for addressing miscalibration within the SSL scenario involves controlling the magnitude of predicted logits for unlabeled samples. Furthermore, from \textbf{Observation 2} we can derive that the underlying mechanism magnifying the miscalibration issue stems from the hidden min-entropy term on the data points in $\mathcal{D}_{U''}$, which represents the majority of unlabeled samples. Thus, %  should prevent the logit magnitudes of unlabeled samples from being large, particularly for incorrect predictions.%Specifically, this strategy should aim to minimize the likelihood of large logit magnitudes, especially when these correspond to incorrect predictions.
%To do so, multiple strategies 
%Even though several strategies have been proposed to adjust the logit values, for example in class imbalance problems \cite{tao2023local}, these do not directly decrease their magnitude, which does not alleviate the miscalibration issue.  
we resort to an inequality constraint that imposes a controllable margin on the logit distances of predictions in samples from $\mathcal{D}_{U''}$. This constraint, which draws inspiration from \cite{liu2022devil}, takes the following form $\dd (\llll) \leq \mathbf{m}$, where $\dd (\llll) = (\max_j (l_j) - l_k)_{1 \leq k \leq K} \in \mathbb{R}^{K} $ represents the vector of logit distances between the winner class and the rest (only for samples in $\mathcal{D}_{U''}$) and $\textbf{m}$ a $K$-dimensional vector defining the margin values, with all elements equal to $m \in \real_{++}$. Note that while \cite{liu2022devil} was proposed in the fully supervised learning scenario, in this work we only enforce the constraint on a subset of samples, which is motivated by the observations presented in \cref{ssec:cal-SSL}. Furthermore, this choice is also supported empirically in the ablation studies presented in the experiments. %l section. 

%\textcolor{red}{Maybe we can highlight here the ratios again, stressing that the third term is the most predominant one, and that enforces uncalibrated predictions, as min-entropy is a lower bound on the Shannon entropy, which much worst gradients.}

By integrating this inequality constraint on the logit distances, training becomes a constrained problem, whose objective can formally defined as:

\begin{equation}
\begin{gathered}
\label{eq:constrained-prob}
    \text{minimize} \quad \mathcal{L}_{\text{T}} \\
    \text{subject to} \quad \dd (\llll) \leq \mathbf{m} \qquad  \mathbf{m} \in \real_{++}, \forall \xx_i \in \mathcal{D}_{U''}.
\end{gathered}
\end{equation}

The above constrained problem in \cref{eq:constrained-prob} can be approximated by penalty-based optimization method, transforming the formulation into an unconstrained problem by using a simple ReLU function: 

%, we resort to an unconstrained approximation using a ReLU function: 

\begin{align}
\label{eq:our-l1}
    \min_{\boldsymbol{\theta}} \quad &  \mathcal{L}_{\text{T}} + \lambda \sum_{i \in \mathcal{D}_{U''}}\sum_k \max(0, \max_j (l_{i,j}) - l_{i,k} - m_k)
\end{align}

\noindent where the second term, i.e., the non-linear ReLU penalty, prevents logit distances from exceeding a predetermined margin $m$, and $\lambda\in\real_+$ is a blending hyperparameter which controls the contribution of the CE loss and the corresponding penalty. The intuition behind this penalty term is simple. For the \textit{winner} logits where the distance with the remaining logits is above the %given 
margin $m$, a gradient will be back-propagated to enforce those values to decrease. As a result, the whole logit magnitudes will decrease, potentially alleviating the miscalibration issue in the set of unlabeled samples $\mathcal{L}_{U''}$, which dominate the SSL training. 

%for the inequality constraint $\dd(\llll)\leq \mm$

%\begin{figure}[ht!]
%\begin{center}   \includegraphics[width=1\linewidth]{images/all_algo_cifar100400.png}
%\end{center}
%   \caption{Ratios of samples with same prediction for weak and strong augmentation with samples that were above the threshold of selection.}
%\label{fig:entropy}
%\end{figure}

\section{Experiments}
%\subsection{Set-up}
\textbf{Datasets.} %For the empirical validation, %To evaluate the proposed approach, 
We resort to the recent Unified Semi-supervised Learning Benchmark for Classification (USB) \cite{wang2022usb}, which compiles a diverse and challenging benchmark across several datasets. In particular, we focus on three popular datasets: \textbf{CIFAR-100} \cite{krizhevsky2010learning}, which has significant value as a standard for fine-grained image classification due to its wide range of classes and detailed object distinctions; \textbf{STL-10} \cite{coates2011analysis}, which is widely recognized for its limited sample size and extensive collection of unlabeled data, rendering it a challenging scenario of special significance in the context of SSL; and \textbf{EuroSAT} \cite{helber2019eurosat}, containing 10 unique fine-grained categories related to earth observation and satellite imagery analysis, and important challenges such as high variability and imbalance classes. %, and noise and artifacts. %of land use and land cover, hence introducing various complications pertinent to the area of earth observation and satellite imagery analysis. 
Last, we also conduct further experiments in the long-tailed version of CIFAR-100.

\noindent \textbf{Architectures.} %In this work, w
We have prioritized Vision Transformers (ViT) over %traditional 
Convolutional Neural Networks (CNNs), for three main reasons related to %\textit{i)} 
discriminative performance, %\textit{ii)} 
quality of uncertainty estimates, %\textit{iii)} 
generalization and transfer learning capabilities. First, the emergence of ViTs has proven these models to outperform their CNNs counterparts \cite{raghu2021do, cai2022semisupervised}. Second, from a calibration standpoint, ViTs have also shown to be better calibrated than CNNs \cite{minderer2021revisiting,pinto2021vision}. Hence, due to their superior discriminative and calibration performance, they pose a more challenging scenario to evaluate the effectiveness of the proposed strategy. And last, the fine-tuning capabilities of ViTs enable effective transfer learning across diverse visual tasks and datasets. This capability is particularly advantageous in scenarios where labeled data is scarce, %or when adapting models to new domains, 
as it allows leveraging pre-trained representations learned from large-scale datasets, significantly minimizing the amount of training iterations while maintaining consistent performance\footnote{Training FreeMatch on CIFAR100 with 400 labeled samples goes from 12 days (WideResNet from scratch) to 10 hours (ViTSmall) in an NVIDIA V100-32G GPU.}. More concretely, we employ a ViTSmall \cite{Gani_2022_BMVC} with a patch size of 2 and an image size of 32 for CIFAR-100 and EuroSAT, in accordance with the standard in USB, and a ViT-Small with an image size of 96 for STL10. %Instead of training ResNets from scratch for CV tasks, we use the pre-trained Vision Transformers (ViT) as it is possible to significantly minimise the amount of training iterations while maintaining consistent performance by employing pre-trained ViTs. We employ a ViTSmall with a patch size of 2 and an image size of 32 for CIFAR-100 and EuroSAT, in accordance with the standard in [USB], and a ViT-Small with an image size of 96 for STL10. % In NLP, we use Bert as the pre-trained backbone. 
We use same settings for all models and benchmarks to provide a fair comparison. %Mention that another motivation of using ViTs is that they are overtaking CNNs, and highlight that it is also more challenging from a calibration perspective, as they are better calibrated (and add refs).

\noindent \textbf{Training, evaluation protocol and metrics.} While several works use different amounts of labeled data in their experiments, we perform due diligence and follow the settings proposed in USB \cite{wang2022usb} for training. For each method and configuration, we perform three runs with different seeds, select the best checkpoint and report their mean and standard deviation, following the literature. We report error rates for the accuracy performance and the expected calibration error (ECE), following the literature in calibration of supervised models. Implementation details are discussed in \appen. 

%Detail the metrics used (Accuracy, ECE, and others, in case we use more). 

\subsection{Results}

\noindent \textbf{Main results}. The proposed strategy is model agnostic, and can be integrated on top of any SSL approach based on pseudo-labels, enabling substantial flexibility. For the empirical evaluation, we selected three popular and relevant approaches that resort to hard pseudo-labels: FixMatch \cite{sohn2020fixmatch}, FlexMatch \cite{zhang2021flexmatch} and FreeMatch \cite{wang2023freematch} and assess the impact of adding our simple solution during training. %on the training of these methods. 
%Following recent literature, we report the mean value and the standard deviation of three random independent runs (each with different seed) of each setting. 
%First, we evaluate the d
\textit{Discriminative performance (\cref{table:tab-main}):} %From results in \cref{table:tab-main}, %whose results are reported in Table \ref{table:tab-main}. In particular, 
we observe that in 16 out of the 18 different settings, adding the penalty in \cref{eq:our-l1} brings improvement gains compared to the original versions of each method, which are only trained with \cref{eq:start-eq}. Note that these gains are sometimes substantial, improving the original method by up to 4\% (e.g., FixMatch in CIFAR-100(200) and EuroSAT(20) or FlexMatch in EuroSAT(20)). Furthermore, the three %baseline 
approaches combined with our penalty achieve very competitive performance compared to existing SSL literature, typically yielding state-of-the-art results. \textit{Calibration performance (\cref{table:tab-main-ECE}):} Similarly, including the penalty term %in \cref{eq:our-l1} %the accuracy of the uncertainty estimations delivered by the methods in \ref{table:tab-main} is reported in Table \ref{table:tab-main-ECE}. 
systematically enhances the calibration performance of the three analyzed approaches, whose improvements are typically significant (up to 6-7\% in several cases). 
%the calibration performance of the compared methods, including FixMatch, Flexmatch and FreeMatch, as well as our modified versions, is reported in Table \ref{table:tab-main-ECE}.
An interesting observation is that, MixMatch \cite{berthelot2019mixmatch}, a consistency regularization based approach, yields surprisingly well calibrated models. Indeed, MixMatch has several components that have shown to improve calibration, such as ensembling predictions and MixUp, which may explain the obtained values. Nevertheless, it is noteworthy to mention that the discriminative performance compared to our modified versions of SSL methods %the SSL methods enhanced with our calibration %our approaches 
is strikingly lower, with differences typically going from 10\% to 24\%, failing to achieve a good compromise between accuracy and calibration. %\textcolor{red}{Furthermore, the reliability plots in \cref{fig:relplots} further bring empirical evidence on how adding the proposed penalty helps to calibrate the original model, particularly for overconfidence situations (above the diagonal).} 

%To explain wht MixMatch could be well calibrated for some cases \cite{thulasidasan2019mixup}.... Furthermore, MixMatch computes average predictions across all augmentations... Indeed MixMatch integrates several components that have shown to improve calibration, such as ensembling predictions and MixUp...

%comment about the calibration performance

\begin{table*}[!t]
\scriptsize
\centering
\caption{\textbf{Classification performance (error rate (\%)).} %for different SSL methods. %across various settings and datasets (CIFAR-100, EuroSAT, and STL-10). 
Arrows indicate whether our modified version improves (\textcolor{blue}{$\downarrow$}) or deteriorates (\textcolor{red}{$\uparrow$}) the performance. Best overall performance in bold and best across pseudo-label SSL approaches underlined.} %Additional approaches are included in Supplemental Material.}% of the original model.}
\label{table:tab-main}
\begin{tabular}{l|P{1.5cm}P{1.5cm}|P{1.5cm}P{1.5cm}|P{1.5cm}P{1.5cm}}
\toprule
Dataset & \multicolumn{2}{c|}{CIFAR-100}& \multicolumn{2}{c|}{EuroSAT} & \multicolumn{2}{c}{STL-10} \\ 
\midrule
\# Labeled samples & 200  & 400  & 20  & 40 & 40  & 100 \\
\midrule
\multicolumn{7}{l}{\textit{\textbf{Only consistency regularization}}} \\
\midrule
%UDA & 27.43 & 18.43 & 13.68 & 4.87 & 15.05 & 9.47 \\
%\midrule
%MeanTeacher & 35.65 & 26.81 & 27.71 & 6.45 & 19.58 & 11.47 \\
%\midrule
MixMatch$_\text{NeurIPS'19}$ & 37.68$_{\pm 2.66}$ & 26.84$_{\pm 1.06}$ & 28.77$_{\pm 10.40}$ & 14.88$_{\pm 2.07}$ & 25.19$_{\pm 2.05}$ & 11.37$_{\pm 1.49}$ \\
%\midrule
Dash$_\text{ ICML'21}$ & 28.51$_{\pm 2.91}$ & 19.54$_{\pm 1.20}$ & 10.05$_{\pm 8.15}$ & 6.83$_{\pm 3.24}$ & 18.30$_{\pm 4.58}$ & 8.74$_{\pm 2.13}$ \\
%\midrule
%CoMatch$_\text{ ICCV'21}$ & 30.26$_{\pm 2.15}$ & 20.94$_{\pm 0.67}$ & 11.90$_{\pm 2.80}$ & 4.75$_{\pm 0.39}$ & 23.17$_{\pm 2.25}$ & 12.76$_{\pm 1.60}$ \\
AdaMatch$_\text{ ICLR'22}$ & \textbf{19.26}$_{\pm 1.83}$ & 17.13$_{\pm 0.92}$ & 12.01$_{\pm 4.16}$ & 6.07$_{\pm 2.26}$ & 13.31$_{\pm 3.75}$ & 8.14$_{\pm 1.48}$ \\
DeFixMatch$_\text{ ICLR'23}$ & 30.44$_{\pm 0.82}$ & 20.93$_{\pm 1.42}$ & 14.27$_{\pm 9.05}$ & 5.42$_{\pm 2.69}$ & 25.36$_{\pm 4.40}$ & 10.97$_{\pm 1.75}$  \\
\midrule
\multicolumn{7}{l}{\textit{\textbf{Pseudo-labeling}}} \\
\midrule
FixMatch$_\text{NeurIPS'20}$ & 31.28$_{\pm 1.58}$ & 19.42$_{\pm 1.56}$ & 11.88$_{\pm 6.32}$ & 6.64$_{\pm 5.03}$ & 16.13$_{\pm 2.36}$ & 8.06$_{\pm 2.15}$ \\
\rowcolor{brightgray} FixMatch + Ours & 27.57$_{\pm 1.49}$\textcolor{blue}{$\downarrow$} & 18.48$_{\pm 1.65}$\textcolor{blue}{$\downarrow$}  & 7.19 $_{\pm 4.83}$\textcolor{blue}{$\downarrow$} & 5.02$_{\pm 2.24}$\textcolor{blue}{$\downarrow$} & 17.55$_{\pm 4.00}$\textcolor{red}{$\uparrow$} & 7.96$_{\pm 1.64}$\textcolor{blue}{$\downarrow$} \\
%\midrule
FlexMatch$_\text{NeurIPS'21}$ & 28.27$_{\pm 0.59}$ & 17.61$_{\pm 0.51}$ & 7.89$_{\pm 3.06}$ & 7.13$_{\pm 1.23}$ & 13.34$_{\pm 1.63}$ & 8.35$_{\pm 1.24}$ \\
\rowcolor{brightgray} FlexMatch + Ours & 26.49$_{\pm 0.52}$\textcolor{blue}{$\downarrow$} & 18.15$_{\pm 0.47}$\textcolor{red}{$\uparrow$}  & \textbf{\underline{3.69}$_{\pm 0.81}$}\textcolor{blue}{$\downarrow$} & 5.00$_{\pm 0.98}$\textcolor{blue}{$\downarrow$} &\textbf{\underline{12.87}$_{\pm 4.32}$}\textcolor{blue}{$\downarrow$} & \textbf{\underline{7.53}$_{\pm 1.32}$}\textcolor{blue}{$\downarrow$} \\
%\midrule
FreeMatch$_\text{ICLR'23}$ & 23.92$_{\pm 2.02}$ & 16.18$_{\pm 0.38}$ & 4.74$_{\pm 1.77}$ & 4.48$_{\pm 0.73}$ & 14.88$_{\pm 0.72}$ & 8.83$_{\pm 0.14}$ \\
\rowcolor{brightgray} FreeMatch + Ours & \underline{21.36}$_{\pm 1.62}$\textcolor{blue}{$\downarrow$} & \textbf{\underline{16.09}$_{\pm 0.80}$}\textcolor{blue}{$\downarrow$}  & 4.30$_{\pm 1.46}$\textcolor{blue}{$\downarrow$} & \textbf{\underline{3.50}$_{\pm 0.70}$}\textcolor{blue}{$\downarrow$} & 13.18$_{\pm 1.61}$\textcolor{blue}{$\downarrow$} & 8.57$_{\pm 1.05}$\textcolor{blue}{$\downarrow$} \\
%\midrule
%SimMatch & 22.49$_{\pm 0.26}$ & 19.45$_{\pm 0.08}$ & 6.24$_{\pm 0.40}$ & 6.63$_{\pm 1.71}$ & 11.3$_{\pm 2.36}$ & 8.90$_{\pm 3.0}$ \\
%SimMatch + Ours & \textbf{23.00$_{\pm 0.07}$}\textcolor{red}{$\uparrow$} & \textbf{19.10$_{\pm 0.31}$}\textcolor{blue}{$\downarrow$}  & \textbf{6.53$_{\pm 0.56}$}\textcolor{red}{$\uparrow$} & \textbf{5.16$_{\pm 1.26}$}\textcolor{blue}{$\downarrow$} & \textbf{11.73$_{\pm 5.15}$}\textcolor{red}{$\uparrow$} & \textbf{8.16$_{\pm 0.57}$}\textcolor{blue}{$\downarrow$} \\
\bottomrule
\end{tabular}
\end{table*}

\begin{table*}[ht!]
\scriptsize
\centering
\caption{\textbf{Calibration performance (ECE).}  %for different SSL methods.} % across various settings and datasets (CIFAR-100, EuroSAT, and STL-10). 
Arrows indicate whether our modified version improves (\textcolor{blue}{$\downarrow$}) or deteriorates (\textcolor{red}{$\uparrow$}) the performance. Best overall performance in bold, whereas best across pseudo-label SSL approaches is underlined.}
\label{table:tab-main-ECE}
\begin{tabular}{l|P{1.5cm}P{1.5cm}|P{1.5cm}P{1.5cm}|P{1.5cm}P{1.5cm}}
\toprule
Dataset & \multicolumn{2}{c|}{CIFAR-100}& \multicolumn{2}{c|}{EuroSAT} & \multicolumn{2}{c}{STL-10} \\ 
\midrule
\# Labeled samples & 200  & 400  & 20  & 40 & 40  & 100 \\
\midrule
\multicolumn{7}{l}{\textit{\textbf{Only consistency regularization}}} \\
\midrule
%UDA & 14.98 & 7.80 & 9.29 & 1.77 & 1.98 & 2.27 \\
%\midrule
%MeanTeacher & 11.17 & 14.10 & 10.50 & 3.38 & 6.12 & 5.88 \\
%\midrule
MixMatch$_\text{NeurIPS'19}$ & \textbf{8.13}$_{\pm 2.16}$ %\mytriangle{blue} 
& \textbf{7.19}$_{\pm 2.31}$ & 9.27$_{\pm 3.68}$ & 3.75$_{\pm 3.30}$ & \textbf{3.42}$_{\pm 1.77}$ & 5.89$_{\pm 1.51}$ \\
%\midrule
Dash$_\text{ ICML'21}$ & 22.23$_{\pm 2.85}$ & 13.20$_{\pm 1.07}$ & 7.09$_{\pm 6.65}$ & 4.24$_{\pm 2.22}$ & 11.23$_{\pm 2.33}$ & 5.23$_{\pm 1.87}$  \\
%\midrule
% CoMatch$_\text{ ICCV'21}$ & 4.91 & 6.97 & 8.07 & 2.54 & 11.51 & 4.62 \\
AdaMatch$_\text{ ICLR'22}$ & 12.96$_{\pm 1.76}$ & 11.17$_{\pm 0.72}$ & 8.55$_{\pm 4.83}$ & 2.66$_{\pm 0.67}$ & 8.80$_{\pm 3.01}$ & 4.96$_{\pm 1.25}$  \\
DeFixMatch$_\text{ ICLR'23}$ & 24.54$_{\pm 1.37}$ & 14.89$_{\pm 0.97}$ & 10.46$_{\pm 8.71}$ & 2.90$_{\pm 1.29}$ & 12.89$_{\pm 1.73}$ & 6.62$_{\pm 1.87}$  \\
\midrule
\multicolumn{7}{l}{\textit{\textbf{Pseudo-labeling}}} \\
\midrule
FixMatch$_\text{NeurIPS'20}$  & 27.77$_{\pm 1.49}$ & 13.45$_{\pm 1.45}$ & 8.36$_{\pm 5.29}$ & 4.72$_{\pm 4.45}$ & 10.27$_{\pm 2.4}$ & 5.83$_{\pm 2.25}$ \\
\rowcolor{brightgray} FixMatch + Ours & 21.56$_{\pm 1.32}$\textcolor{blue}{$\downarrow$} & 12.12$_{\pm 1.70}$\textcolor{blue}{$\downarrow$}  & 4.66$_{\pm 2.49}$\textcolor{blue}{$\downarrow$} & 3.84$_{\pm 1.91}$\textcolor{blue}{$\downarrow$} & 7.83$_{\pm 4.23}$\textcolor{blue}{$\downarrow$} & 5.64$_{\pm 1.40}$\textcolor{blue}{$\downarrow$} \\
%\midrule
FlexMatch$_\text{NeurIPS'21}$ & 21.95$_{\pm 0.57}$ & 11.95$_{\pm 0.30}$ & 5.42$_{\pm 2.95}$ & 4.50$_{\pm 2.60}$ & 9.72$_{\pm 1.63}$ & 5.85$_{\pm 0.98}$ \\
\rowcolor{brightgray} FlexMatch + Ours & 19.74$_{\pm 0.50}$\textcolor{blue}{$\downarrow$} & 11.61$_{\pm 0.29}$\textcolor{blue}{$\downarrow$}  & \textbf{\underline{2.26}}$_{\pm 1.13}$\textcolor{blue}{$\downarrow$} & 3.21$_{\pm 1.79}$\textcolor{blue}{$\downarrow$} & 8.89$_{\pm 3.39}$\textcolor{blue}{$\downarrow$} & 4.97$_{\pm 1.19}$\textcolor{blue}{$\downarrow$} \\
%\midrule 
FreeMatch$_\text{ICLR'23}$ & 18.27$_{\pm 1.60}$ & 11.56$_{\pm 0.44}$ & 3.49$_{\pm 1.39}$ & 3.22$_{\pm 0.55}$ & 10.49$_{\pm 1.87}$ & 5.24$_{\pm 0.90}$ \\
\rowcolor{brightgray} FreeMatch + Ours & \underline{14.86}$_{\pm 1.22}$\textcolor{blue}{$\downarrow$} & \underline{10.35}$_{\pm 0.68}$\textcolor{blue}{$\downarrow$}  & 2.82$_{\pm 0.81}$\textcolor{blue}{$\downarrow$} & \textbf{\underline{2.63}}$_{\pm 0.70}$\textcolor{blue}{$\downarrow$} & \underline{3.74}$_{\pm 0.90}$\textcolor{blue}{$\downarrow$} & \textbf{\underline{3.50}}$_{\pm 1.02}$\textcolor{blue}{$\downarrow$} \\
%\midrule
%SimMatch & 3.17$_{\pm 0.28}$ & 6.54$_{\pm 1.92}$ & 1.72$_{\pm 0.17}$ & 3.78$_{\pm 0.24}$ & 1.97$_{\pm 1.84}$ & 2.43$_{\pm 0.64}$ \\
%SimMatch + Ours & \textbf{3.31$_{\pm 0.59}$}\textcolor{red}{$\uparrow$} & \textbf{5.55$_{\pm 0.13}$}\textcolor{blue}{$\downarrow$}  & \textbf{2.09 $_{\pm 0.07}$}\textcolor{red}{$\uparrow$} & \textbf{2.93$_{\pm 1.91}$}\textcolor{blue}{$\downarrow$} & \textbf{2.93$_{\pm 3.70}$}\textcolor{red}{$\uparrow$} & \textbf{2.01$_{\pm 0.70}$}\textcolor{blue}{$\downarrow$} \\
\bottomrule
\end{tabular}
\end{table*}

We would like to stress that our goal is not to propose a novel state-of-the-art SSL approach, but to shed light about an important issue in prevalent pseudo-labeling based SSL methods, and present a simple yet efficient solution that can improve their performance. These results %reported 
%in these tables 
demonstrate that integrating the penalty in \cref{eq:our-l1} during training appears as an appealing strategy to improve both accuracy %error rate 
and calibration performance 
of pseudo-label SSL approaches. %Reliability plots are depicted in Supplemetal Material. 

\begin{figure}[ht!]
    \centering
    \includegraphics[width=0.65\linewidth]{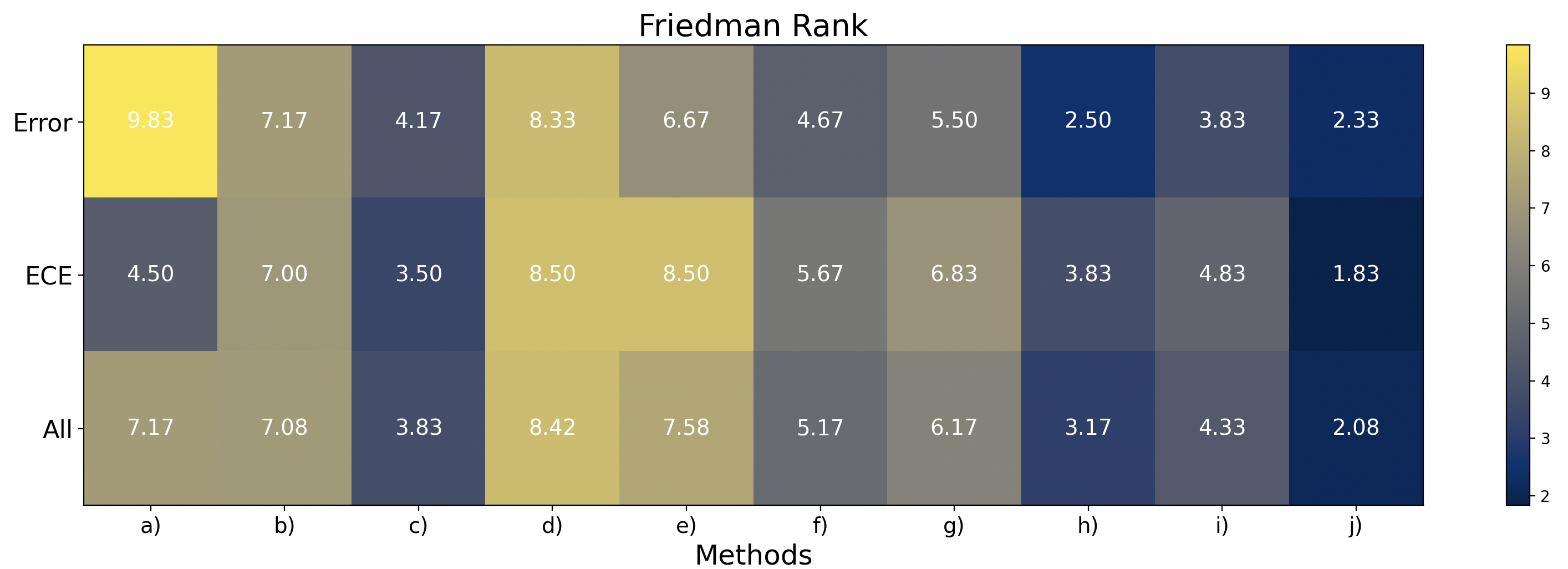}
    \caption{\textbf{Friedman Rank} for the methods analyzed in Tables \ref{table:tab-main} and \ref{table:tab-main-ECE} following \cite{wang2022usb}: \textbf{a)} MixMatch, \textbf{b)} Dash, \textbf{c)} AdaMatch, \textbf{d)} DeFixMatch, \textbf{e)} Fixmatch, \textbf{f)} FixMatch + Ours, \textbf{g)} FlexMatch, \textbf{h)} FlexMatch + Ours, \textbf{i)} FreeMatch and \textbf{j)} FreeMatch + Ours.} %When analyzing three different scenarios (i.e., Error only, ECE and both) integrating our simple penalty yields the best performing models across overall.}
    \label{fig:friedman-rank}
\end{figure}

Last, we follow \cite{wang2022usb} and employ the Friedman rank \cite{friedman1937use,friedman1940comparison} to fairly compare the performance of different methods across various settings. This metric can be defined as $\operatorname{rank}_F=\frac{1}{m} \sum_{i=1}^m \operatorname{rank}_i$, with $m$ being the number of evaluation settings ($m=12$ in our case, 2 metrics $\times$ 6 datasets), and $\operatorname{rank}_i$ the rank of a method in the \textit{i}-th setting. Hence, the lower the rank obtained, the better the method. These rankings, which are depicted in \cref{fig:friedman-rank}, show that our modified versions provide competitive performances considering accuracy (Error), ECE and both (All), with our FreeMatch and FlexMatch versions presenting the best alternatives across overall methods.

%We further show in Figure \ref{fig:relplots} the reliability plots.. The confidence histogram at the bottom shows how many test examples are in each bin...The two vertical lines indicate the overall accuracy and average confidence. The closer these two lines are together, the better the model is calibrated.

%\begin{figure*}[ht!]
%         \centering
%    \includegraphics[width=\linewidth]{images/STL10.pdf}
%     \caption{Reliability plots : STL10 (a) Top Row : Methods (b) Bottom Row : Method + Ours} 
%     \label{fig:relplots2}
%\end{figure*}

\begin{figure}[ht!]
\centering
    \includegraphics[width=\textwidth]{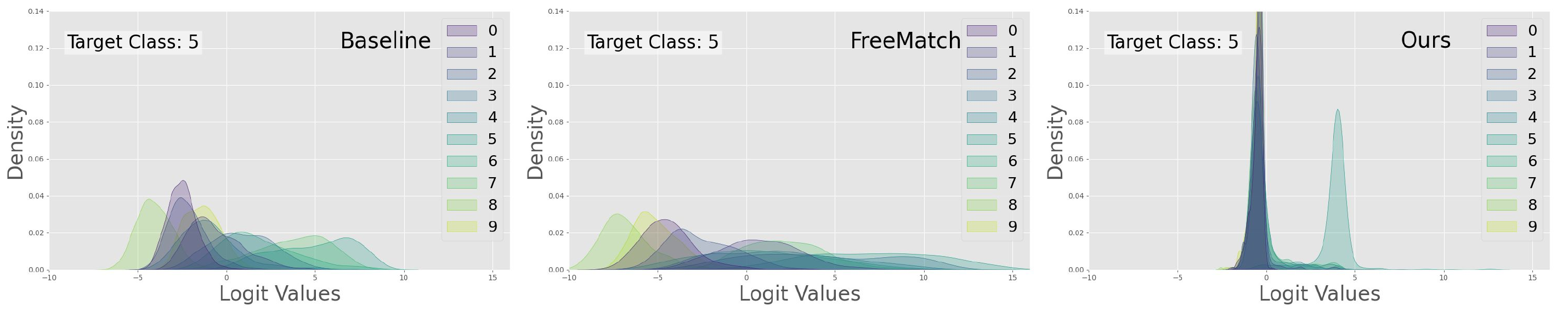}
    \caption{\textbf{Impact of the proposed solution in the logits}, which plots the Kernel density estimation of the logits distribution (per-class) for target class 5 for the supervised baseline (\textit{left}), original FreeMatch (\textit{middle}) and our version (\textit{right}). }
    \label{fig:logitsDistResults}
    %\caption{In-depth analysis of the effect of the proposed solution. We show the distribution of the logits magnitude after training. Study case on STL-10 with 40 labeled samples (4 per class).}
\end{figure}

\noindent \textbf{On the impact on the logits.} %distribution}. %In addition to the main results shown in Tables \ref{table:tab-main} and \ref{table:tab-main-ECE}, w
We now analyze in more detail the impact of incorporating the constraint during training, %(\cref{eq:our-l1}), 
particularly on the logit distribution. %To this end, we select FixMatch as SSL approach and the STL-10 with 40 labeled samples as study-case, due to the discriminative performance degradation observed in our method. First, we can observe in Figures \ref{fig:image1} and \ref{fig:image2} that, despite yielding a lower classification error compared to our modified version (16.13\% \textit{vs.} 17.55\%), it typically brings significant larger confusion across several classes (e.g., categories 3,4 and 5). Indeed, if we look at the per-class error analysis... HERE SHOW ERROR PER CLASS... Furthermore, as depicted in \ref{fig:image3}, 
In \cref{fig:logitsDistResults}, we depict the kernel density estimation of the logits distribution, which was negatively affected by SSL methods compared to a fully-supervised baseline (\textit{Observation 3}). From these figures, we can observe two interesting findings that will have a positive effect in calibration: \textit{i)} the logit magnitude of the incorrect predicted classes is decreased, and \textit{ii)} the whole logit range also decreases. This means that, even for incorrect predictions, the reduced logit values will lead to less confident predictions, which contrasts with the potentially more confidence scores obtained by SSL methods. We argue that the reduction of the logit magnitudes is a byproduct of % We argue that this is due to 
the penalty preventing the logit differences to be large. This limits the effect of the min-entropy pushing hard towards the vertex of the simplex, which is minimized with %and therefore yielding 
either 0 or 1 softmax predictions. %, and therefore the logits also see their magnitudes decrease. %i.e., softmax probability to either 0 or 1...performances observed..... 
%bla bla bla, 
Thus, following our hypothesis from Observation 3, we can advocate that our strategy provides well-calibrated SSL methods, as both the total range and logit magnitude of incorrect predictions are significantly decreased. 
%We show here the case of STL 10. We show \ref{fig:four_images} as a study case (maybe in one class), and then we can show these distributions for all datasets.

%In this section we will try to analyze in some detail the 

\begin{figure*}[ht!]
         \centering
    \includegraphics[width=\linewidth]{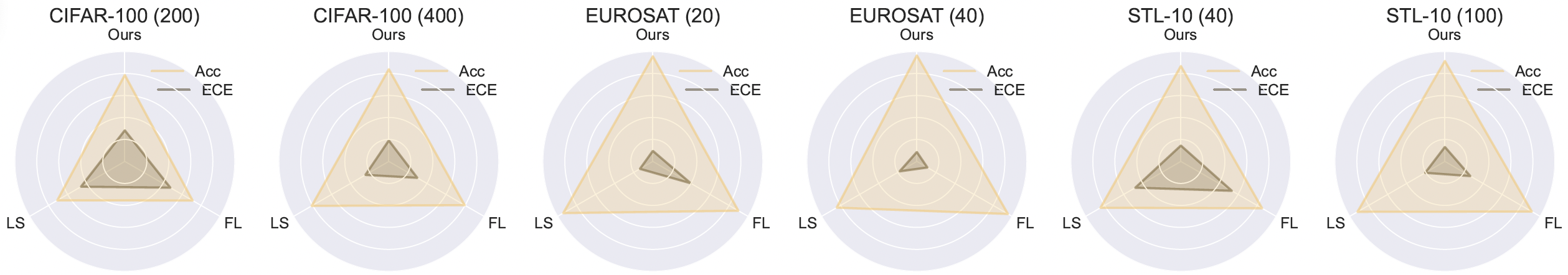}
     \caption{Radar plots for different calibration approaches based on FreeMatch. The range of the radar plot for ECE is changed for better visualization (CIFAR-100: [1, 50], EuroSAT: [1, 20], STL-10: [1, 20])} 
     \label{fig:radar}
\end{figure*}

\noindent \textbf{Comparison to other calibration methods}. In this section, we compare our approach to relevant %state-of-the-art 
calibration approaches %, which have been proposed 
in the fully-supervised scenario. In particular, we evaluate the impact of adding label smoothing (LS) \cite{szegedy2016rethinking} and focal loss (FL) \cite{lin2017focal} % and adaptive and conditional Label Smoothing (ACLS) \cite{park2023acls} 
to FreeMatch, as it is the most recent studied method. %among the three studied in this work. 
Results from these experiments, depicted %as radar plots 
in \cref{fig:radar}, confirm that our approach consistently yields the best discriminative-calibration trade-off across datasets and settings, which can be measured by the largest gap between the accuracy (\textit{bright yellow}) and ECE (\textit{dark yellow}). %\textbf{Comparison with BAM}. 
Furthermore, we compare to BAM \cite{loh2023mitigating}, up to our knowledge the only concurrent work that tackles calibration work on SSL. This comparison (\cref{tab-BAM}) demonstrates that our simple solution outperforms the recent BAM in terms of error rate and ECE, emerging as a promising choice.

%... BAM is a competing approach that adds a Bayesian layer for the calibration of the model. We compare it with our simple calibration technique in Tab. \ref{tab-BAM} on CIFAR100 for 200 and 400 annotated samples. In both cases our approach improves over BAM in terms of error of both rate and ECE.

%\textit{Full-supervised} (Focal loss, Label smoothing) and \textit{semi-supervised (BAM)}. 

\begin{table*}[ht!]
\scriptsize
\centering
\caption{Discriminative and calibration performance compared to BAM \cite{loh2023mitigating}.}% Best results in bold.}
\label{tab-BAM}
\begin{tabular}{l|P{1.5cm}P{1.5cm}|P{1.5cm}P{1.5cm}}
\toprule
Method & \multicolumn{2}{c|}{CIFAR100 (200)} & \multicolumn{2}{c}{CIFAR100 (400)} \\
\midrule
& Error & ECE & Error & ECE  \\
\midrule
FixMatch + BAM & 28.46$_{\pm 1.74}$ & 25.04$_{\pm 1.43}$  & $19.32_{\pm 0.94}$ & 16.75$_{\pm 0.96}$\\
\rowcolor{brightgray} FixMatch + Ours & $\textbf{27.57}_{\pm 1.49}$ & $\textbf{21.56}_{\pm 1.32}$ & $\textbf{18.42}_{\pm 1.65}$ & $\textbf{12.12}_{\pm 1.70}$\\

\bottomrule
\end{tabular}
\end{table*}

%\noindent \textbf{Ablation on terms in which the penalty is added}.
\noindent \textbf{In which samples to impose the penalty?} A natural question that arises from our analysis is whether constraining the logit distances should also be applied to the samples in $\mathcal{D}_{U'}$. We argue that, indeed, it is only beneficial to the samples in $\mathcal{D}_{U''}$. While we formally exposed that the third term in \cref{eq:our-l1} minimizes the min-entropy, the second term has some sort of corrective effect, where pseudo-labels from weak augmentations correct the predictions obtained with the strong augmented versions. As labels are different, enforcing a penalty on the logit distances might indeed have a counterproductive effect. More concretely, trying to satisfy the constraint in samples from $\mathcal{D}_{U'}$, where hard predictions %pseudo-labels (or predictions) 
are from different classes, may actually impede the network to learn semantically meaningful features that can bring discriminative capabilities to the model. %... \textcolor{red}{(thinking about this)..}. 
To support these arguments empirically, we report in \cref{table:terms} the performance of the three SSL methods - FixMatch, FlexMatch and FreeMatch when the constraint is enforced across different scenarios. %only in the last term (i.e., our proposed model) and in both unlabeled terms. 
We can clearly see that, regardless of the configuration or model used, including the penalty on the samples from $\mathcal{D}_{U'}$ has a detrimental effect, with discriminative and calibration results suffering a substantial degradation.

\begin{table}[ht!]
    \scriptsize
    \centering
    \caption{\textbf{Impact of the penalty term} in the different unlabeled subsets, $\mathcal{D_{U'}}$ and $\mathcal{D_{U''}}$ for the three relevant pseudo-labeling SSL methods studied in this work. The proposed approach is shadowed in gray and best result in bold. }
    \label{table:terms}
    \begin{tabular}{@{}lcccccc@{}}
        \toprule
         & & & \multicolumn{2}{c|}{CIFAR-100 (200)} &  \multicolumn{2}{c}{CIFAR-100 (400)}  \\
        \midrule
        & $\mathcal{D}_{U'}$ & $\mathcal{D}_{U''}$ &  Error & ECE & Error & ECE \\
        \midrule
        \multirow{3}{*}{FixMatch} &  & & 27.77$_{\pm 1.49}$ & 31.28$_{\pm 1.58}$ & 15.76$_{\pm 0.39}$ & 19.73$_{\pm 0.56}$  \\
      &  & \ding{51} &  21.56$_{\pm 1.32}$ &  27.57$_{\pm 1.49}$ & 13.49$_{\pm 0.80}$ & \textbf{18.57}$_{\pm 0.77}$ \\
          &  \ding{51} & \ding{51} & \CC{100} \textbf{19.30}$_{\pm 2.52}$ &  \CC{100} \textbf{25.72}$_{\pm 2.36}$ & \CC{100} \textbf{13.00}$_{\pm 1.79}$ & \CC{100} 19.20$_{\pm 1.63}$\\
        % Different Pseudo-labels & & 20.98$_{\pm 1.77}$ & 27.48$_{\pm 1.68}$ \\

        \midrule
        \multirow{3}{*}{FlexMatch} &  & &21.95$_{\pm 0.57}$ & 28.27$_{\pm 0.59}$ & 11.95$_{\pm 0.30}$ & \textbf{17.61}$_{\pm 0.51}$  \\
         &  & \ding{51}  & \CC{100} \textbf{19.74}$_{\pm 0.50}$ & \CC{100} \textbf{26.49}$_{\pm 0.52}$ & \CC{100}\textbf{11.61}$_{\pm 0.29}$ & \CC{100} 18.15$_{\pm 0.47}$ \\

         %Different Pseudo-labels & & 11.83$_{\pm 0.73}$ & 18.95$_{\pm 1.03}$ \\
         & \ding{51}  & \ding{51} &  20.98$_{\pm 3.05}$ & 27.57$_{\pm 3.37}$ &  13.41$_{\pm 0.51}$ & 19.62$_{\pm 0.44}$  \\
        \midrule
       \multirow{3}{*}{FreeMatch} &  & & 23.92$_{\pm 2.02}$ & 18.27$_{\pm 1.95}$ & 16.18$_{\pm 0.38}$ & 11.56$_{\pm 0.53}$    \\
        &  & \ding{51}  & \CC{100}  \textbf{21.36}$_{\pm 1.62}$ & \CC{100} \textbf{14.86}$_{\pm 1.48}$ & \CC{100}  \textbf{16.09}$_{\pm 0.80}$ & \CC{100}  \textbf{10.35}$_{\pm 0.83}$  \\
       & \ding{51}  & \ding{51} & 27.44$_{\pm 1.08}$ & 20.86$_{\pm 1.48}$ & 18.82$_{\pm 0.67}$ & 12.39$_{\pm 0.78}$ \\
        \bottomrule
    \end{tabular}
\end{table}

%\noindent \textbf{Further results on related, but not similar approaches.}

\noindent \textbf{Long-tailed experiments}. 
In \cref{tab-LT} we report error rate and ECE on CIFAR100-LT %, across the three SSL approaches, 
to asses the effect of the proposed calibration when the class population is imbalanced.
%We evaluate FixMatch with different values of $\gamma_l$ and $\gamma_u$. 
Across all cases both the method accuracy and the ECE improve consistently. It is noteworthy to mention that for all these experiments we use the same margin as in CIFAR-100, %a standard margin of 10, 
without any adaptation to the specific setting. 
%Stress that we did  not fine-tune the margin in this setting, but use the same as in the standard SSL experiments... Results are reported in Table \ref{tab-LT}.

\begin{table*}[ht!]
\scriptsize
\centering
\caption{Qualitative performance on long-tailed classification (CIFAR-100-LT).}
\label{tab-LT}
\begin{tabular}{l|P{1.5cm}P{1.5cm}|P{1.5cm}P{1.5cm}|P{1.5cm}P{1.5cm}}
\toprule
Method & \multicolumn{2}{c|}{$\gamma_l$ = 10, $\gamma_u$ = -10} & \multicolumn{2}{c|}{$\gamma_l$ = 10, $\gamma_u$ = 10} & \multicolumn{2}{c}{$\gamma_l$ = 15, $\gamma_u$ = 15} \\
\midrule
& Error & ECE & Error & ECE & Error & ECE \\
\midrule
FixMatch & $15.05_{\pm 0.13}$ & $8.58_{\pm 0.82}$ & $15.01_{\pm 0.18}$ & $10.51_{\pm 1.82}$ & $16.45_{\pm 0.11}$ & $10.22_{\pm 2.01}$ \\
\rowcolor{brightgray} FixMatch + Ours & $\textbf{14.73}_{\pm 0.12}$ & $\textbf{7.22}_{\pm 1.20}$ & $\textbf{14.37}_{\pm 0.21}$ & $\textbf{9.42}_{\pm 0.29}$ & $\textbf{15.81}_{\pm 0.31}$ & $\textbf{8.44}_{\pm 0.44}$ \\
\midrule
FlexMatch & $14.79_{\pm 0.28}$ & $7.70_{\pm 1.03}$ & $14.98_{\pm 0.21}$ & $9.49_{\pm 0.47}$ & $16.19_{\pm 0.32}$ & $9.25_{\pm 0.68}$ \\
\rowcolor{brightgray} FlexMatch + Ours & $\textbf{14.72}_{\pm 0.23}$ & $\textbf{7.33}_{\pm 1.49}$ & $\textbf{14.6}_{\pm 0.05}$ & $\textbf{7.44}_{\pm 0.40}$ & $\textbf{16.08}_{\pm 0.49}$ & $\textbf{8.85}_{\pm 0.47}$ \\
\midrule
FreeMatch & $14.93_{\pm 0.20}$ & $8.52_{\pm 0.95}$ & $15.04_{\pm 0.09}$ & $9.99_{\pm 0.52}$ & $16.26_{\pm 0.35}$ & $8.68_{\pm 0.84}$ \\
\rowcolor{brightgray} FreeMatch + Ours & $\textbf{14.67}_{\pm 0.28}$ & $\textbf{6.74}_{\pm 0.98}$ & $\textbf{14.77}_{\pm 0.15}$ & $\textbf{9.05}_{\pm 0.84}$ & $\textbf{15.87}_{\pm 0.69}$ & $\textbf{8.53}_{\pm 0.60}$ \\
\bottomrule
\end{tabular}
\end{table*}

\noindent \textbf{Additional experiments} with detailed analysis across methods, comparison to additional approaches and additional settings are reported in \appen.

\section{Conclusion}
In this work we have raised awareness of the miscalibratiion problem induced by pseudo-label training, which is one of the most popular approaches for SSL. We demonstrated that the unsupervised loss used by pseudo-label methods is dominated by the min-entropy term, a lower bound of the Shannon entropy, and identified it as a potential source of miscalibration. We then proposed a simple solution based on enforcing a fixed margin constraint between the winner class and its contenders. 
Our solution on popular %commonly used 
SSL datasets and methods %typically 
yields consistent calibration improvements, whereas %. %in most of the cases. 
%As a byproduct, as the considered methods depend on the quality of selected pseudo-labels, %are based on selecting confident samples, 
we also found consistent gains in terms of predictive accuracy, typically %in some cases even 
outperforming the state-of-the-art for SSL.

\section*{Acknowledgments}

This work was funded by the Natural Sciences and Engineering Research Council of Canada (NSERC). We also thank Calcul Quebec and Compute Canada. 

%The paper ends with a conclusion. 

%\clearpage  % TODO REVIEW/FINAL: This \clearpage needs to be removed from both review and camera-ready versions.

\bibliographystyle{splncs04}
\bibliography{main}

\clearpage

%\begin{document}

\newcommand{\red}[1]{\textcolor{red}{#1}}

\resetlinenumber
\setcounter{page}{1}
\setcounter{section}{0}
%\setcounter{table}{0}
%\setcounter{figure}{0}
%\section*{Supplemental Materials. Submission $\#$3472}

\section*{\centering \Large Supplementary Material for \\
Do not trust what you trust: \\ Miscalibration in Semi-supervised Learning}

%\maketitle
\begin{comment}
\section{Code}

In order to reproduce the results reported in this work, we share our code in the following repository : \\
\href{https://anonymous.4open.science/r/Semi-supervised-learning-3525}{https://anonymous.4open.science/r/Semi-supervised-learning-3525} 
\\
The code will be made publicly available so that the research community can benefit from this work, which we expect can be established as a strong baseline for calibrating pseudo-label semi-supervised learning (SSL) approaches.
\end{comment}
\section{Additional discriminative results}

\begin{table*}[!h]
\scriptsize
\centering
\caption{Comparison of error rate (\%) for different SSL methods across various labeled settings and datasets (CIFAR-100, EuroSAT, and STL-10).$\dagger$ indicates that the results are reported from \cite{wang2022usb}. Best results in bold, whereas second-best are underlined.}
\label{table:tab-main-supp}
\begin{tabular}{l|P{1.4cm}P{1.4cm}|P{1.4cm}P{1.4cm}|P{1.4cm}P{1.4cm}}
\toprule
Dataset & \multicolumn{2}{c|}{CIFAR-100}& \multicolumn{2}{c|}{EuroSAT} & \multicolumn{2}{c}{STL-10} \\ 
\midrule
\# Labeled samples & 200  & 400  & 20  & 40 & 40  & 100 \\
\midrule
%UDA & 27.43 & 18.43 & 13.68 & 4.87 & 15.05 & 9.47 \\
%\midrule
%MeanTeacher & 35.65 & 26.81 & 27.71 & 6.45 & 19.58 & 11.47 \\
%\midrule
%\midrule

%\midrule
%CoMatch & 31.21 & 20.76 & 12.79 & 4.88 & 21.12 & 13.9 \\
$\pi$-Model$_\text{ NeurIPS'15}$ $\dagger$&  36.06$_{\pm0.15}$ & 26.52$_{\pm0.41}$	& 21.82$_{\pm1.22}$	&12.09$_{\pm2.27}$ &42.76$_{\pm15.94}$ &	19.85$_{\pm13.02}$	 \\
Mean-Teacher$_\text{ NeurIPS'17}$ & 35.47$_{\pm 0.40}$ & 26.03$_{\pm 0.30}$ & 26.83$_{\pm 1.46}$ & 15.85$_{\pm 1.66}$ & 18.67$_{\pm 2.66}$ & 24.19$_{\pm 10.15}$ \\
VAT$_\text{ TPAMI'18}$ $\dagger$&  31.49$_{\pm1.33}$	&21.34$_{\pm0.50}$ &	18.45$_{\pm1.47}$	&26.16$_{\pm0.96}$	& 10.09$_{\pm0.94}$ & 10.69$_{\pm0.51}$	   \\
MixMatch$_\text{ NeurIPS'19}$ & 37.68$_{\pm 2.66}$ & 26.84$_{\pm 1.06}$ & 28.77$_{\pm 10.40}$ & 14.88$_{\pm 2.07}$ & 25.19$_{\pm 2.05}$ & 11.37$_{\pm 1.49}$ \\
ReMixMatch$_\text{ ICLR'20}$$\dagger$ & 22.21$_{\pm2.21}$ &	16.86$_{\pm0.57}$	&5.05$_{\pm1.05}$	&5.07$_{\pm0.56}$  &13.08$_{\pm3.34}$&	\textbf{7.21}$_{\pm0.39}$	  \\
UDA$_\text{ NeurIPS'20}$ & 28.80$_{\pm 0.61}$ & 19.00$_{\pm 0.79}$ & 9.83$_{\pm 2.15}$ & 6.22$_{\pm 1.36}$ & 15.58$_{\pm 3.16}$ & 7.65$_{\pm 1.11}$ \\
CRMatch$_\text{ GCPR'21}$$\dagger$ & 25.70$_{\pm1.75}$	&18.03$_{\pm0.20}$&	13.24$_{\pm1.69}$	&8.35$_{\pm1.71}$ & \textbf{10.17}$_{\pm0.00}$&	-- 	 \\
CoMatch$_\text{ ICCV'21}$$\dagger$&35.08$_{\pm0.69}$&	25.35$_{\pm0.50}$ &5.75$_{\pm0.43}$	&4.81$_{\pm1.05}$	&15.12$_{\pm1.88}$	&9.56$_{\pm1.35}$	 \\
Dash$_\text{ ICML'21}$ & 28.51$_{\pm 2.91}$ & 19.54$_{\pm 1.20}$ & 10.05$_{\pm 8.15}$ & 6.83$_{\pm 3.24}$ & 18.30$_{\pm 4.58}$ & 8.74$_{\pm 2.13}$ \\
%\midrule
%Dash & 30.61$_{\pm 0.98}$ & 19.38$_{\pm 0.10}$ & 11.19$_{\pm 0.90}$ & 6.96$_{\pm 0.87}$ & 16.22$_{\pm 5.95}$ & 7.85$_{\pm 0.74}$ \\
%AdaMatch$_\text{ ICLR'22}$ $\dagger$ & 22.32$_{\pm1.73}$ & 16.66$_{\pm0.62}$ &  7.02$_{\pm0.79}$	& 4.75$_{\pm1.10}$ & 13.64$_{\pm2.49}$&	7.62$_{\pm1.90}$ \\
AdaMatch$_\text{ ICLR'22}$ & \textbf{19.26}$_{\pm 1.83}$ & 17.13$_{\pm 0.92}$ & 12.01$_{\pm 4.16}$ & 6.07$_{\pm 2.26}$ & 13.31$_{\pm 3.75}$ & 8.14$_{\pm 1.48}$ \\
SimMatch$_\text{ CVPR'22}$ $\dagger$ & 23.78$_{\pm1.08}$&17.06$_{\pm0.78}$ &  7.66$_{\pm0.60}$	& 5.27$_{\pm0.89}$ & \underline{11.77}$_{\pm3.20}$&	7.55$_{\pm1.86}$ \\
SoftMatch$_\text{ ICLR'23}$ $\dagger$ & 22.67$_{\pm1.32}$&16.84$_{\pm0.66}$ &  5.75$_{\pm0.62}$	& 5.90$_{\pm1.42}$ & 13.55$_{\pm3.16}$&	7.84$_{\pm1.72}$ \\
%DefixMatch$_\text{ ICLR'23}$ $\dagger$ & 31.52$_{\pm1.85}$&21.12$_{\pm1.74}$ &  14.71$_{\pm6.52}$	& \underline{3.72}$_{\pm0.79}$ & 17.68$_{\pm7.94}$&	7.94$_{\pm1.31}$ \\
DeFixMatch$_\text{ ICLR'23}$ & 30.44$_{\pm 0.82}$ & 20.93$_{\pm 1.42}$ & 14.27$_{\pm 9.05}$ & 5.42$_{\pm 2.69}$ & 25.36$_{\pm 4.40}$ & 10.97$_{\pm 1.75}$  \\

\midrule
FixMatch$_\text{ NeurIPS'20}$ & 31.28$_{\pm 1.58}$ & 19.73$_{\pm 0.56}$ & 11.88$_{\pm 6.32}$ & 6.64$_{\pm 5.03}$ & 16.13$_{\pm 2.36}$ & 8.06$_{\pm 2.15}$ \\
\rowcolor{brightgray} FixMatch + Ours & 27.57$_{\pm 1.49}$ & 18.57$_{\pm 0.77}$ & 7.19 $_{\pm 4.83}$ & 5.02$_{\pm 2.24}$ & 17.55$_{\pm 4.00}$ & 7.96$_{\pm 1.64}$\\
%\midrule
FlexMatch$_\text{ NeurIPS'21}$ & 28.27$_{\pm 0.59}$ & 17.61$_{\pm 0.51}$ & 7.89$_{\pm 3.06}$ & 7.13$_{\pm 1.23}$ & 13.34$_{\pm 1.63}$ & 8.35$_{\pm 1.24}$ \\
\rowcolor{brightgray} FlexMatch + Ours & 26.49$_{\pm 0.52}$ & 18.15$_{\pm 0.47}$ & \textbf{3.69$_{\pm 0.81}$} & 5.00$_{\pm 0.98}$ & 12.87$_{\pm 4.32}$ & \underline{7.53}$_{\pm 1.32}$ \\
%\midrule
FreeMatch$_\text{ ICLR'23}$ & 23.92$_{\pm 2.02}$ & \underline{16.18}$_{\pm 0.38}$ & 4.74$_{\pm 1.77}$ & 4.48$_{\pm 0.73}$ & 14.88$_{\pm 0.72}$ & 8.83$_{\pm 0.14}$ \\
\rowcolor{brightgray} FreeMatch + Ours & \underline{21.36}$_{\pm 1.62}$ &\textbf{16.09}$_{\pm 0.80}$ & \underline{4.30}$_{\pm 1.46}$ & \textbf{3.50$_{\pm 0.70}$} & 13.18$_{\pm 1.61}$ & 8.57$_{\pm 1.05}$\\
\bottomrule
\end{tabular}
\end{table*}

In the main paper, we included several relevant SSL methods to compare both error and calibration performance. We did not include a considerably large set of approaches because we need to run three times each method in each setting (so that each evaluated method means 3$\times$6=18 runs, with an average of 10 hours per run). Furthermore, we would like to stress again that our goal in this work is not to provide a novel state-of-the-art SSL method, but investigate the miscalibration issue of pseudo-labeling SSL, identify the source of the problem, and provide a solution that can enhance the calibration of this family of methods. Having said this, the discriminative performance of additional methods is provided in \cite{wang2022usb}, which is used to complement the discriminative comparison in Table \ref{table:tab-main}, whose results can be found in Table \ref{table:tab-main-supp} of this appendix.

\section{Can we add our calibration strategy to other methods?}

In the main paper we have evaluated the impact of incorporating the proposed term into relevant SSL approaches that resort to hard pseudo-labels during training. Indeed, our motivation stems from the observations that under this scenario (i.e., the standard pseudo-labeling process in SSL), there is a hidden min-entropy objective that dominates the training. While we have demonstrated that our approach effectively improves both the discriminative and calibration performance of these models, we further assess its impact on approaches that use soft pseudo-labels, i.e., pseudo-labels on weak augmentations are not converted to one-hot encoded vectors. For these approaches, the reasoning in \cref{eq:tot1} and \cref{eq:tot2} does not entirely hold in these cases. More concretely, the recent work SimMatch \cite{zheng2022simmatch} uses this strategy\footnote{Note that most pseudo-labeling SSL approaches use a hard pseudo-label strategy, where the softmax predictions of weak augmentations are transformed into one-hot encoded pseudo-labels.}, where the direct softmax prediction over the weak augmented samples is used as supervisory signal for its strong augmented counterpart. The results from this experiment are reported in Table \ref{table:tab-simmatch}, which show that adding the proposed penalty has a positive effect favouring better calibrated models even in methods that use soft pseudo-labels, particularly as the number of labeled samples increases. Note that, as our main findings, observations, and motivations, do not entirely apply for soft pseudo-labels approaches, such as SimMatch, we do not make any claim regarding the benefits of our strategy in these approaches. In contrast, we simply show empirically that even in these cases, our method can still bring performance benefits.

%\red{Here ECE augments with the number of labeled samples...this is weird...}

\begin{table*}[!h]
\scriptsize
\centering
\caption{Comparison to a related approach, SimMatch, which uses soft pseudo-labels, instead of hard pseudo-labels (as FixMatch, FlexMatch and FreeMatch). }
\label{table:tab-simmatch}
\begin{tabular}{l|P{1.5cm}P{1.5cm}P{1.5cm}|P{1.5cm}P{1.5cm}P{1.5cm}}
\toprule
& \multicolumn{3}{c}{Error rate (\%)} & \multicolumn{3}{c}{ECE}   \\
\midrule
  & \multicolumn{6}{c}{CIFAR-100}   \\ 
\midrule
\# Labeled samples & 200  & 400 & 1000  & 200  & 400 & 1000 \\
\midrule
SimMatch & \textbf{22.49}$_{\pm 0.26}$ & 19.45$_{\pm 0.08}$ & 15.46$_{\pm 0.48}$ & \textbf{3.17}$_{\pm 0.28}$ & 6.54$_{\pm 1.92}$ & 9.60$_{\pm 0.72}$\\
\rowcolor{brightgray} SimMatch + Ours & 23.00$_{\pm 0.07}$\textcolor{red}{$\uparrow$} & \textbf{19.10$_{\pm 0.31}$}\textcolor{blue}{$\downarrow$}  & \textbf{15.36}$_{\pm 0.41}$\textcolor{blue}{$\downarrow$} & 3.31$_{\pm 0.59}$\textcolor{red}{$\uparrow$} & \textbf{5.55$_{\pm 0.13}$}\textcolor{blue}{$\downarrow$} & \textbf{9.18}$_{\pm 0.17}$\textcolor{blue}{$\downarrow$} \\
%\midrule
% & \multicolumn{6}{c}{EuroSAT}  \\ 
% \midrule
% \# Labeled samples & 20  & 40 & 100  & 20  & 40 & 100 \\
%\midrule
%SimMatch &  \textbf{6.24}$_{\pm 0.40}$ & 6.63$_{\pm 1.71}$ & 5.08$_{\pm 2.39}$ & \textbf{1.72}$_{\pm 0.17}$ & 3.78$_{\pm 0.24}$ & 3.91$_{\pm 2.00}$ \\
%\rowcolor{brightgray} SimMatch + Ours & 6.53$_{\pm 0.56}$\textcolor{red}{$\uparrow$} & \textbf{5.16$_{\pm 1.26}$}\textcolor{blue}{$\downarrow$} & \textbf{4.55}$_{\pm 1.75}$\textcolor{blue}{$\downarrow$} & 2.09 $_{\pm 0.07}$\textcolor{red}{$\uparrow$} & \textbf{2.93$_{\pm 1.91}$}\textcolor{blue}{$\downarrow$} & \textbf{3.33}$_{\pm 1.59}$\textcolor{blue}{$\downarrow$}  \\
%\midrule
% & \multicolumn{6}{c}{STL-10} \\ 
%\midrule
%\# Labeled samples & 20  & 40 & 100  & 20  & 40 & 100 \\
%\midrule
%SimMatch &  \textbf{11.30}$_{\pm 2.36}$ & 8.90$_{\pm 3.0}$ & 4.42$_{\pm 0.18}$ & \textbf{1.97}$_{\pm 1.84}$ & 2.43$_{\pm 0.64}$ & 1.90$_{\pm 0.62}$\\
%\rowcolor{brightgray} SimMatch + Ours & 11.73$_{\pm 5.15}$\textcolor{red}{$\uparrow$} & \textbf{8.16$_{\pm 0.57}$}\textcolor{blue}{$\downarrow$} & 4.57$_{\pm 0.25}$\textcolor{red}{$\uparrow$}& 2.93$_{\pm 3.70}$\textcolor{red}{$\uparrow$} & \textbf{2.01$_{\pm 0.70}$}\textcolor{blue}{$\downarrow$} & 2.08$_{\pm 0.31}$\textcolor{red}{$\uparrow$}\\
\bottomrule
\end{tabular}
\end{table*}

\section{Dataset details}

We provide detailed insights into the datasets utilized in our experiments, whose configuration is strongly inspired by the recent USB (A Unified Semi-supervised Learning Benchmark for Classification) benchmark \cite{wang2022usb}. For \textbf{CIFAR-100}, a renowned benchmark for fine-grained image classification, we considered two label settings: 2 labeled samples and 4 labeled samples per class for each of the 100 classes,resulting in a total of 50,000 training samples and 10,000 samples for testing. Each image in CIFAR-100 is sized at 32×32 pixels. \textbf{STL-10}, known for its limited sample size and extensive unlabeled data, offers a unique challenge. We employed two label settings as well: 4 labeled samples and 10 labeled samples per class for all 10 classes, %\red{with 5,000 labeled samples} 
and an additional 100,000 unlabeled samples for training, along with 8,000 samples for testing. Each STL-10 image measures 96×96 pixels. Lastly, \textbf{EuroSAT}, based on Sentinel-2 satellite images, features two label settings: 2 labeled samples per class and 4 samples per class for 10 classes. With a total of 16,200 training samples, including labeled and unlabeled images and 5,400 testing samples, EuroSAT images are sized at 64×64. While literature includes SVHN and CIFAR-10 datasets, we do not use them for evaluation, as state-of-the-art SSL methods already achieve a performance comparable to that of fully supervised training on these datasets. The selected datasets offer diverse challenges and settings, allowing for a comprehensive evaluation of semi-supervised learning methods across various domains.

For the \textbf{long-tailed CIFAR-100} experiments, we investigate two subsets of the dataset, with $N1$ representing the number of samples in the minority class and $M$ denoting the number of samples in the majority class. The imbalance ratio is controlled by $\gamma_l$ and $\gamma_u$, where both parameters are set to -10, 10 or 15, ensuring a consistent level of label imbalance across categories. In the first setting, we set the imbalance ratio of labeled samples $\gamma_l$ to 10 and the imbalance ratio of unlabeled samples $\gamma_u$ to 10, with the number of labeled samples (N1) set to 150 and the number of unlabeled samples (M) set to 300. In the second setting, we maintain the same label imbalance ratio ($\gamma_l$ = 10) and number of labeled samples (N1 = 150), but we set the unlabeled sample imbalance ratio $\gamma_u$ to -10. Finally, in the third setting, we increase the $\gamma_l$ to 15 while keeping the $\gamma_u$ at 15, with N1 and M set to 150 and 300, respectively. This configuration introduces a higher degree of label imbalance, with the minority class having 15 times fewer samples than the majority class. We follow the same setup as authors in \cite{wang2022usb} for the classic setting using WideResnet. These settings allow us to comprehensively evaluate the performance of machine learning models under varying degrees of label imbalance and class distribution scenarios within the CIFAR100-LT dataset.

\section{Implementation details}

%We include CIFAR-100 [39] and STL-10 [40] from TorchSSL since they are still challenging

%Here all the hyperparameters used, explanations about different methods also using different hyperparameters depending on each setting. Ideally, some ablation to motivate some choices (e.g. LR on BAM, or even marging for us (if it plays in our favour)). 

%we follow \cite{wang2022usb,zhang2021flexmatch} to report the best number of all checkpoints to avoid unfair comparisons caused by different convergence speeds. 

%At some point, talk about the slowest convergence of BAM..

%from BAM: The only exception is the sharpening temperature of BaM-UDA, which uses t = 0.9 instead of t = 0.4 in UDA...

%ACLS has 4 hyperparam - we use same margin for them and for us..
\noindent \textit{Method-dependent hyperparemeters.}
In the context of pseudo-labeling methods, while FixMatch and FlexMatch rely on a single threshold hyperparameter for pseudo-label selection, FreeMatch introduces additional hyperparameters such as a pre-defined threshold $\tau$, unlabeled batch ratio $\mu$, unsupervised loss weight $w_u$, fairness loss weight $w_f$, and EMA decay $\lambda$. To avoid unfair comparisons across methods, we use the default values for all these hyperparameters, which are reported in their respective papers. %Notably, all other algorithm-specific hyperparameters remained consistent with their original specifications as detailed in the respective papers.

\noindent \textit{Specific-case: BAM.} In our comparison with BAM \cite{loh2023mitigating}, we encountered difficulties in hyperparameter tuning. BAM employs the quantile $Q$ over the batch to determine the threshold for pseudo-label selection, which serves as a key hyperparameter. In the cited paper, $Q = 0.75$ for the CIFAR-100 benchmark and $Q = 0.95$ for the CIFAR-10 benchmark were chosen. Additionally, BAM incorporates a separate Adam optimizer for the Bayesian Neural Network (BNN) layer, with a fixed learning rate of 0.01. Notably, for the sharpening temperature parameter in BaM-UDA, BAM opts for $t = 0.9$ as opposed to $t = 0.4$ utilized in UDA. These hyperparameter selections are specific to both the dataset and the method employed, thus presenting a challenge when evaluating the method across different datasets. For our experiments, we chose to focus on CIFAR-100 due to the inherent complexity involved in conducting extensive hyperparameter searches, not only concerning the dataset or method, but also regarding the optimizer and learning rate for the introduced BNN layer.

\noindent \textit{Margin Selection with Limited Fine-tuning:} We stress that in the majority of cases, we did not extensively fine-tune the margin hyperparameter. Across various experiments, settings and datasets, the chosen margin values remained consistent without significant adjustments. Specifically, for the CIFAR-100 and EuroSAT datasets, the margins were uniformly set without requiring further refinement, with values of 10 and 8 across methods, respectively. Given its complexity, we needed to perform a hyperparameter search in a few cases in the STL-10 dataset. In particular, most settings employed a margin equal to 10, similar to the CIFAR-100 dataset. Nevertheless, we observe that, particularly for FreeMatch, this margin did not yield the best performance across the two settings of labeled data (i.e., 40 and 100 labeled samples). After fine-tuning the margin value, we found that for STL-10 (40) and STL-10 (100) FreeMatch worked best with a margin of 4 and 6, respectively. %Even in scenarios where hyperparameter search was conducted, such as with the STL-10 dataset for the FixMatch algorithm, the margin adjustments were limited, with the margin values remaining relatively stable across different labeled data scenarios --7 for 40 labeled instances and 10 for 100 labeled instances--. For FlexMatch, a constant margin of 10 was utilized across both scenarios, reflecting its consistency in margin selection. Similarly, FreeMatch employed specific margin values depending on the number of labeled instances: a margin of 4 for scenarios with 40 labeled instances and a margin of 6 for scenarios with 100 labeled instances. 
Furthermore, for all the remaining experiments with CIFAR-100, as well as with long-tailed CIFAR-100, we kept the margin fixed ($m=10$) across experiments and methods.

\noindent \textit{Training hyperparemeters.} Regarding algorithm-independent hyperparameters, we adhered to the settings outlined in \cite{wang2022usb}. Specifically, the learning rate was set to $5\times10^{-4}$ for CIFAR-100, $10^{-4}$ for STL-10, and $5\times10^{-5}$ for EuroSAT. During training, the batch size was fixed at 8, while for evaluation, it was set to 16. Additionally, the layer decay rate varied across datasets: 0.5 for CIFAR-100, 0.95 for STL-10, and 1.0 for EuroSAT. Weak augmentation techniques employed included random crop and random horizontal flip, while strong augmentation utilized RandAugment \cite{cubuk2020randaugment}. The cosine annealing scheduler was utilized with a total of 204,800 steps and a warm-up period of 5,120 steps. Both labeled and unlabeled batch sizes were set to 16. Furthermore, we follow \cite{wang2022usb,zhang2021flexmatch} to report the best number of all checkpoints to avoid unfair comparisons caused by different convergence speeds.

\section{Model choices}

In this work we have selected three relevant SSL approaches based on hard pseudo-labels, as our findings are closely related to these approaches. More concretely, we demonstrate empirically our observations in FixMatch \cite{sohn2020fixmatch}, FlexMatch \cite{zhang2021flexmatch} and FreeMatch \cite{wang2023freematch}, as they are popular SSL methods, some of them published recently (i.e., FreeMatch has been published in 2023). Thus, for the Table \ref{table:tab-main} and Table \ref{table:tab-main-ECE}, we have shown the effect of adding the proposed term in all the three approaches. Similarly, the impact of our term over the three methods is evaluated in Table \ref{tab-LT} (i.e., CIFAR-100-LT), as we consider that long-tailed classification represents an important task, and assessing the performance of the three approaches will undoubtedly strengthen the message that our solution can improve both classification and calibration performance across a general family of SSL approaches. Regarding the ablation studies to evaluate the impact of several choices, we have used FreeMatch, the most recent approach among the three, in all the ablations (Fig \ref{fig:radar} and Table \ref{table:terms}), except when comparing to the concurrent work in BAM \cite{loh2023mitigating}. The reason is that authors in \cite{loh2023mitigating} only consider FixMatch in their experiments, despite BAM being a work published in 2023. Thus, the code\footnote{\href{https://github.com/clott3/BaM-SSL}{https://github.com/clott3/BaM-SSL}} is all constrained to the use of FixMatch and CIFAR-100, making it difficult to evaluate with other SSL approaches and datasets. In fact, we attempted to integrate FreeMatch on the BAM framework, but the results obtained were suboptimal. We believe that, the sensitivity to hyperparameters (e.g., just the choice of the backbone for the Bayesian layer in BAM requires different learning rates) could be the main cause for the low performance of BAM in SSL approaches other than FixMatch. Therefore, in order to avoid an unfair comparison (due to the suboptimal performance of BAM in other SSL methods) we decided to compare to it based only on FixMatch, following their original work.

\section{Numerical values for results in radar plots}

For the radar plots shown in the main paper, we present here (Table \ref{tab:fl_and_ls}) the individual quantitative results, for a more detailed comparison across methods. As a reminder, FreeMatch \cite{wang2023freematch} is used as a baseline strategy, and the different calibration strategies (i.e., Focal Loss \cite{lin2017focal}, Label Smoothing \cite{szegedy2016rethinking}, and ours) are applied over the $\mathcal{D}_{U''}$ subset.

\begin{table}[ht!]
\scriptsize
\centering
\caption{Numerical values of radar plots depicted in Figure 6 of the main paper.}
\label{tab:fl_and_ls}
\begin{tabular}{@{}lcccccc@{}}
\toprule
Method & Setting & Error & ECE & CECE & AECE \\
\midrule
Baseline & \multirow{4}{*}{CIFAR 100 (200)} & 23.92$_{\pm 2.02}$ & 18.27$_{\pm 1.95}$ & 0.41$_{\pm 0.04}$ & 18.27$_{\pm 1.95}$ \\
Focal Loss & & 28.81$_{\pm 0.75}$ & 24.42$_{\pm 0.85}$ & 0.53$\pm_{0.01}$ & 24.42$_{\pm 0.85}$ \\
Label Smoothing & & 29.14$_{\pm 2.46}$ & 23.55$_{\pm 2.33}$ & 0.52$_{\pm 0.04}$ & 23.54$_{\pm 2.34}$ \\
\rowcolor{brightgray} \textbf{Ours} & & \textbf{21.36$_{\pm 1.62}$} & \textbf{14.86$_{\pm 1.48}$} & \textbf{0.35$_{\pm 0.03}$} & \textbf{14.81$_{\pm 1.48}$} \\
\midrule
Baseline & \multirow{4}{*}{CIFAR 100 (400)} & 16.18$_{\pm 0.38}$ & 11.56$_{\pm 0.53}$ & 0.27$_{\pm 0.01}$ & 11.54$_{\pm 0.52}$ \\
Focal Loss & & 20.35$_{\pm 0.35}$ & 15.63$_{\pm 0.54}$ & 0.35$_{\pm 0.008}$ & 15.60$_{\pm 0.54}$ \\
Label Smoothing & & 19.01$_{\pm 1.90}$ & 13.03$_{\pm 1.39}$ & 0.31$_{\pm 0.02}$ & 13.02$_{\pm 1.39}$ \\
\rowcolor{brightgray} \textbf{Ours} & & \textbf{16.09$_{\pm 0.80}$} & \textbf{10.35$_{\pm 0.83}$} & \textbf{0.26$_{\pm 0.01}$} & \textbf{10.33$_{\pm 0.83}$} \\
\midrule
Baseline & \multirow{4}{*}{EuroSAT (20)} & 4.74$_{\pm 1.82}$ & 3.50$_{\pm 1.70}$ & 0.82$_{\pm 0.32}$ & 3.38$_{\pm 1.82}$ \\
Focal Loss & & 10.25$_{\pm 8.48}$ & 8.54$_{\pm 8.29}$ & 1.88$\pm_{1.70}$ & 8.52$_{\pm 8.31}$ \\
Label Smoothing & & 5.94$_{\pm 4.21}$ & 3.57$_{\pm 2.83}$ & 0.95$_{\pm 0.67}$ & 3.45$_{\pm 2.81}$ \\
\rowcolor{brightgray} \textbf{Ours} & & \textbf{4.30$_{\pm 1.46}$} & \textbf{2.82$_{\pm 1.00}$} & \textbf{0.70$_{\pm 0.23}$} & \textbf{2.78$_{\pm 0.98}$} \\
\midrule
Baseline & \multirow{4}{*}{EuroSAT (40)} & 4.48$_{\pm 0.73}$ & 3.22$_{\pm 0.55}$ & 0.74$_{\pm 0.11}$ & 3.15$_{\pm 0.61}$ \\
Focal Loss & & 4.37$_{\pm 0.30}$ & 3.10$_{\pm 0.10}$ & 0.74$\pm_{0.04}$ & 3.01$_{\pm 0.08}$ \\
Label Smoothing & & 15.84$_{\pm 3.57}$ & 4.43$_{\pm 3.38}$ & 1.02$_{\pm 0.70}$ & 4.37$_{\pm 3.31}$ \\
\rowcolor{brightgray} \textbf{Ours} & & \textbf{3.50$_{\pm 0.70}$} & \textbf{2.63}$_{\pm 0.70}$ & \textbf{0.60}$_{\pm 0.12}$ & \textbf{2.58}$_{\pm 0.64}$ \\
\midrule
Baseline & \multirow{4}{*}{STL 10 (40)} & 14.88$_{\pm 0.72}$ & 10.49$_{\pm 1.87}$ & 2.45$_{\pm 0.40}$ & 9.40$_{\pm 1.81}$ \\
Focal Loss & & 14.73$_{\pm 2.77}$ & 11.16$_{\pm 3.13}$ & 2.39$\pm{0.64}$ & 11.15$_{\pm 3.13}$ \\
Label Smoothing & & \textbf{15.59$_{\pm 2.74}$} & 10.09$_{\pm 1.64}$ & \textbf{2.20$_{\pm 0.28}$} & 10.04$_{\pm 1.58}$ \\
\rowcolor{brightgray} \textbf{Ours} & & \textbf{13.18$_{\pm 1.61}$} & \textbf{3.74$_{\pm 5.64}$} & 2.39$_{\pm 0.68}$ & \textbf{8.43$_{\pm 4.92}$} \\
\midrule
Baseline & \multirow{4}{*}{STL 10 (100)} & 7.62$_{\pm 1.25}$ & 5.24$_{\pm 1.09}$ & 1.15$_{\pm 0.21}$ & 5.19$_{\pm 1.03}$ \\
Focal Loss & & 8.75$_{\pm 1.59}$ & 6.07$_{\pm 1.40}$ & 1.35$_\pm{0.31}$ & 6.02$_{\pm 1.35}$ \\
Label Smoothing & & \textbf{8.25$_{\pm 2.05}$} & 4.94$_{\pm 1.79}$ & 1.13$_{\pm 0.35}$ & 4.94$_{\pm 1.79}$ \\
\rowcolor{brightgray} \textbf{Ours} & & 8.57$_{\pm 1.90}$ & \textbf{3.50$_{\pm 1.54}$} & \textbf{0.94$_{\pm 0.27}$} & \textbf{4.00$_{\pm 1.21}$} \\
\bottomrule
\end{tabular}
\end{table}

\section{Detailed comparison between original and our versions}
We include additional comparisons between our proposed method and the original SSL approaches. 

\begin{figure*}[ht!]
         \centering
    \includegraphics[width=0.9\linewidth]{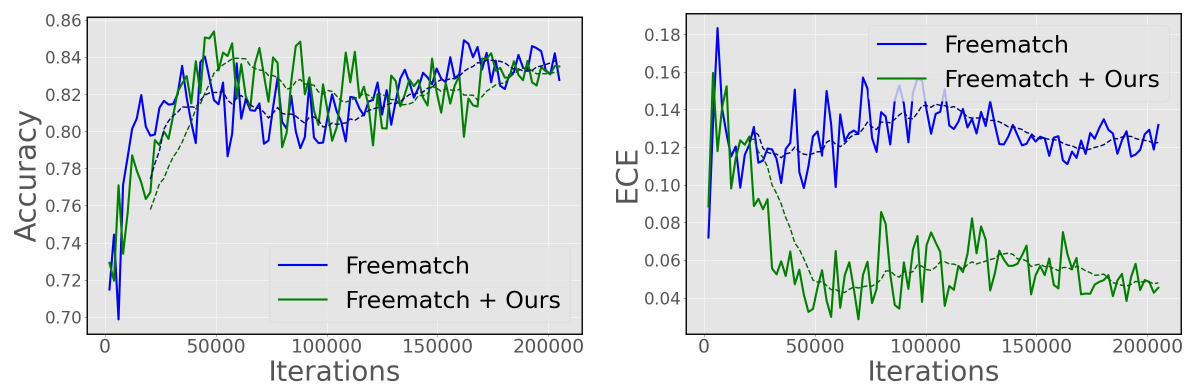}
     \caption{Comparison of convergence of Accuracy (\textit{left}) and ECE (\textit{right}) over iterations for STL10 dataset with 40 labeled samples.} 
     \label{fig:convergence}
\end{figure*}

\noindent \textit{Convergence analysis.} First, as depicted in Fig. \ref{fig:convergence}, we assess the impact of adding our approach to FreeMatch in terms of convergence. %of the different algorithms. %the performance of the algorithms is evaluated across iterations to assess their convergence. 
Notably, our method, denoted as FreeMatch (Ours), demonstrates superior performance in terms of accuracy, outperforming the baseline FreeMatch method within the initial $50,000$ iterations. In contrast, the original method reaches convergence around $170,000$ iterations, making it considerably slower if one looks at discriminative performance. Additionally, our approach consistently achieves lower values of Expected Calibration Error (ECE) throughout the iterative process, highlighting its potential to enhance the calibration performance. This contrasts with the original FreeMatch, whose predictions seem to be better calibrated at the first iteration, and these are degraded (i.e., ECE increases) with the training. Thus, adding our method exhibits accelerated convergence rates compared to the original approach, reaching competitive performance levels in a fraction of the time.

%Reliability plots, accuracy plots and logit distributions for all (or several) settings... Add also the convergence plots and pseudo-labels quality...

\begin{figure*}[ht!]
         \centering
    \includegraphics[width=0.9\linewidth]{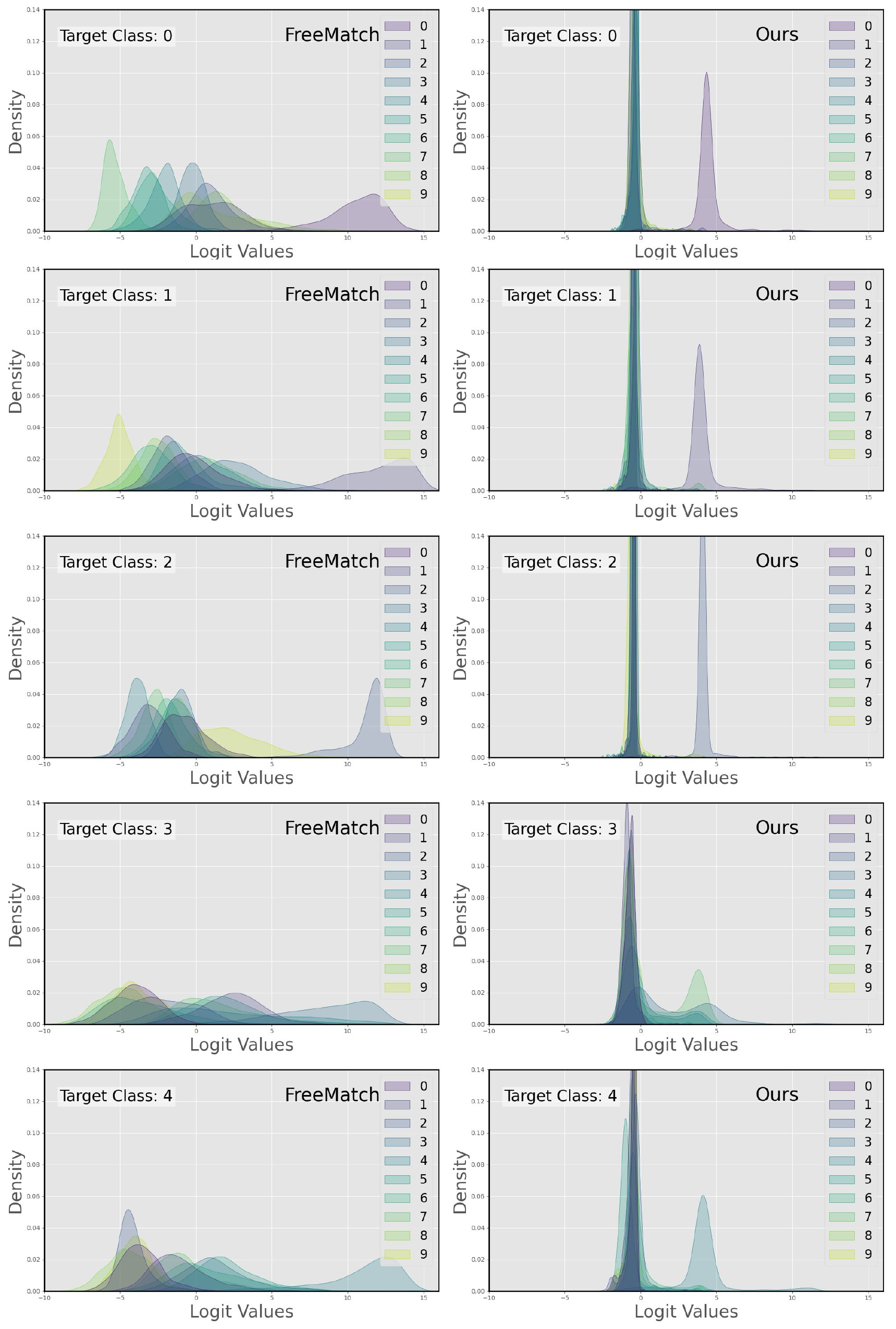}
     \caption{Comparison of convergence of Accuracy (\textit{left}) and ECE (\textit{right}) over iterations for STL10 dataset with 40 labeled samples.} 
     \label{fig:free_vs_ours_0_4}
\end{figure*}

\begin{figure*}[ht!]
         \centering
    \includegraphics[width=0.9\linewidth]{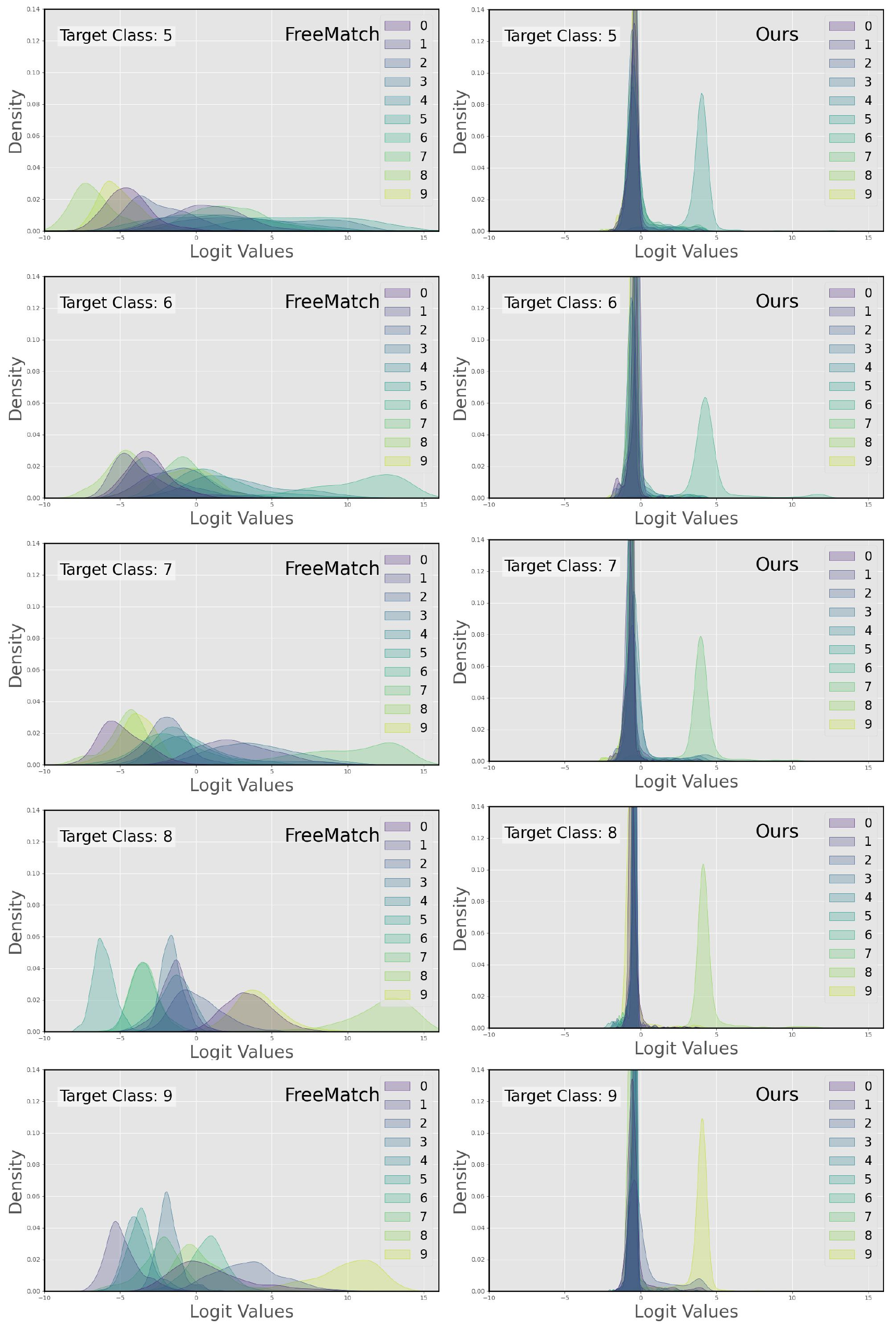}
     \caption{Comparison of convergence of Accuracy (\textit{left}) and ECE (\textit{right}) over iterations for STL10 dataset with 40 labeled samples.} 
     \label{fig:free_vs_ours_5_9}
\end{figure*}

\noindent \textit{Logits distribution.} We now provide additional insights into the effects of our approach across the logit distribution. More concretely, in \cref{fig:free_vs_ours_0_4} and \cref{fig:free_vs_ours_5_9} we present the kernel density estimation of the logits distribution for each class on the STL10 dataset for FreeMatch and FreeMatch plus our method (denoted as Ours). %of semi-supervised learning (SSL) methods on the calibration of the model across all 10 classes of the STL10 dataset for FreeMatch and our method. In Fig. \ref{fig:free_vs_ours_0_4}, we present the kernel density estimation of the logits distribution for each class for FreeMatch, comparing SSL methods against ours in Fig. \ref{fig:free_vs_ours_5_9}. 
This analysis expands on Observation 3 from the main paper, highlighting the impact of SSL on the calibration of the model across a diverse range of classes. Specifically, while the original FreeMatch provides larger logit ranges, as well as larger logit values even for incorrect classes, adding our approach limits these logit increases across all the classes. As the logit range decreases, the resulting softmax probabilities for the predicted classes will be smaller, alleviating the problem of overconfident predictions. This is particularly important for wrong predictions, as we expect a well-calibrated model to not be overconfident when incorrect classes are predicted.  
%we observe a decrease in the logit magnitude of incorrectly predicted classes, as well as a reduction in the overall logit range for our method as compared to the original method. 

%\begin{figure}[htbp]
%    \centering
%    \begin{minipage}[b]{0.45\linewidth}
%        \centering
%        \foreach \i in {0,1,2,3,4}{
%            \begin{subfigure}[b]{0.9\linewidth}
%                \centering
%                \includegraphics[width=\linewidth]{images/freematch_stl10_40_ours/kde_logit_distribution_class_\i.pdf}
%                \caption{Class \i}
%                \label{fig:class\i}
%            \end{subfigure}
%        }
%    \end{minipage}%
%   \begin{minipage}[b]{0.45\linewidth}
%       \centering
%       \foreach \i in {5,6,7,8,9}{
%           \begin{subfigure}[b]{0.9\linewidth}
%               \centering
%               \includegraphics[width=\linewidth]{images/freematch_stl10_40_ours/kde_logit_distribution_class_\i.pdf}
%                \caption{Class \i}
%                \label{fig:class\i}
%            \end{subfigure}
%        }
%    \end{minipage}
%    \caption{FreeMatch : Ours STL 10 Classwise KDE Plots}
%    \label{fig:freematch_ours_classwise_kde}
%\end{figure}

\begin{figure*}[ht!]
         \centering
    \includegraphics[width=\linewidth]{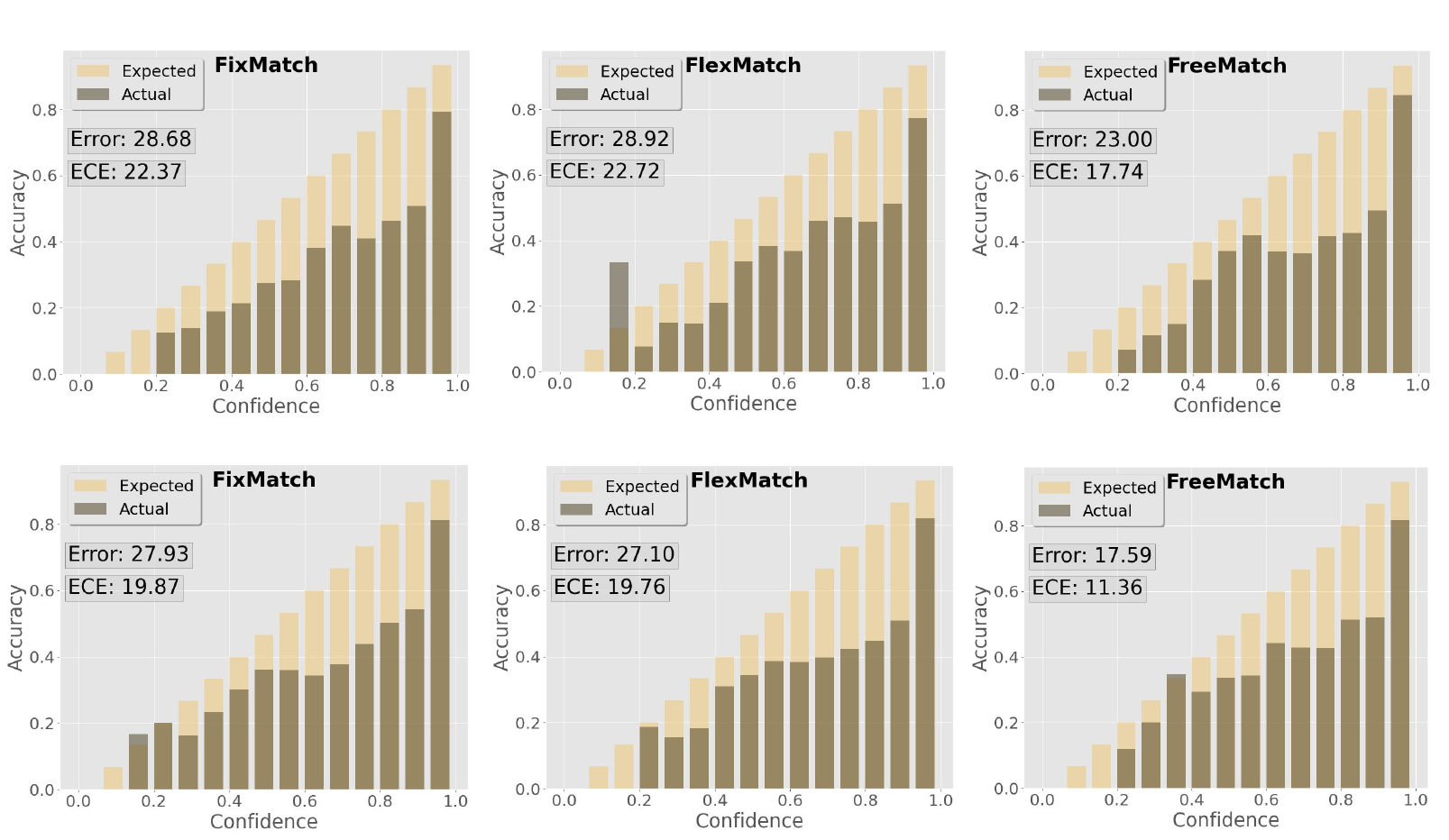}
     \caption{Reliability plots on CIFAR-100(200) for the original (\textit{top}) and our (\textit{bottom}) versions of the three pseudo-label SSL methods selected.} 
     \label{fig:relplots}
\end{figure*}

\begin{figure*}[ht!]
         \centering
    \includegraphics[width=\linewidth]{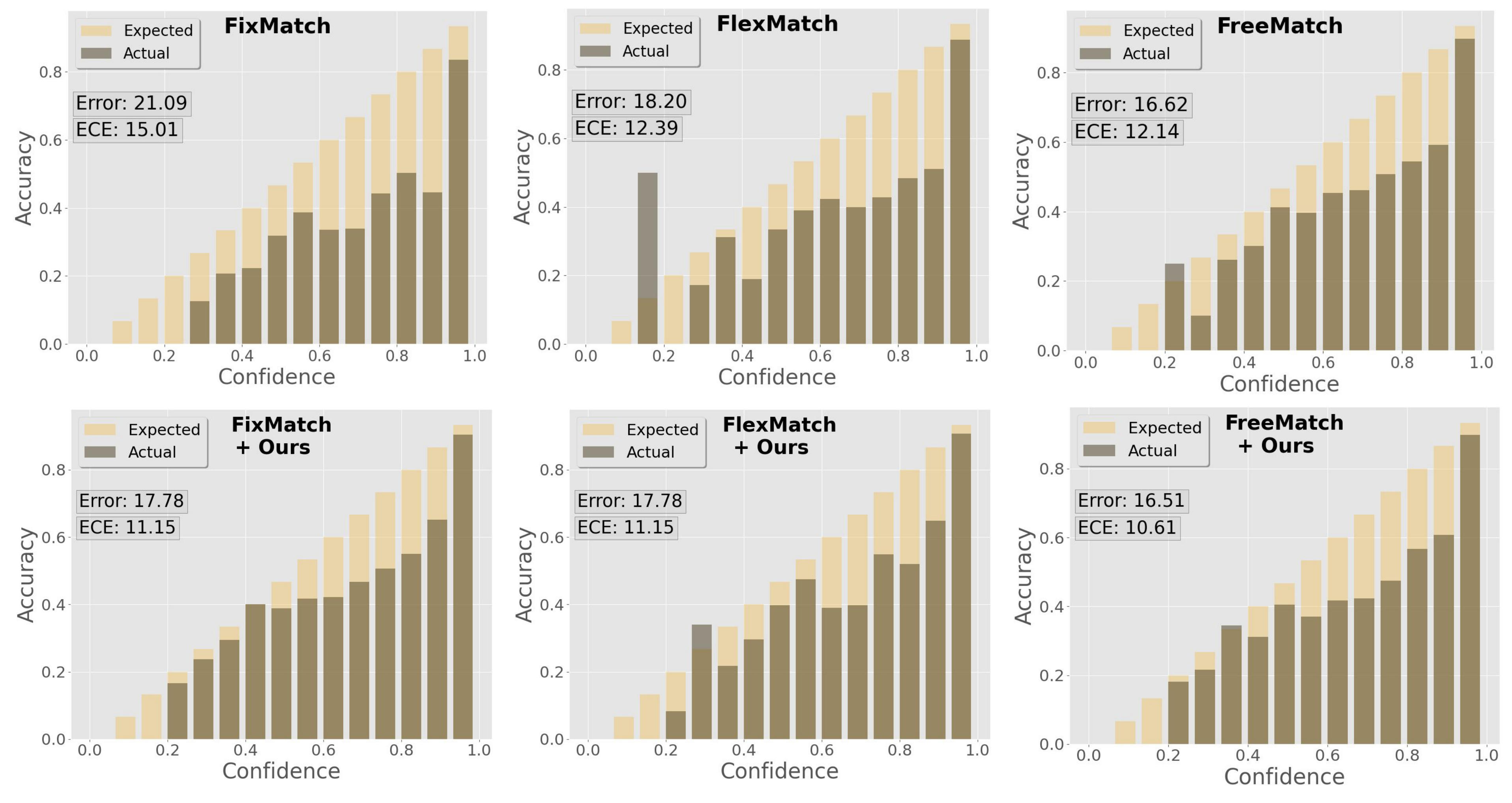}
     \caption{Reliability plots on CIFAR-100(400) for the original (\textit{top}) and our (\textit{bottom}) versions of the three pseudo-label SSL methods selected.} 
     \label{fig:ece_acc_plots}
\end{figure*}

\noindent \textit{Reliability plots.} Last, we depict several reliability plots for both the semi-supervised learning (SSL) methods that were discussed in the main paper and our modified version, to highlight the calibration improvements brought by our approach (\cref{fig:relplots} and \cref{fig:ece_acc_plots}).

%By introducing additional reliability plots for the semi-supervised learning (SSL) methods that were discussed in the main paper (\cref{fig:relplots} ), we further elaborate on the findings presented therein. Additionally, we also include the reliability plots for the proposed method, when it is integrated in the three representative SSL approaches studied in this work. A main observation that motivated our work was the decline in calibration of pseudo-label SSL approaches (Figure 1 of main paper)

%Additionally, we present our proposed method. In contrast to the main paper, which predominantly compared three SSL methods with a supervised baseline in the supplementary graphs, these now also include our proposed method in conjunction with the selected SSL techniques. By utilising the knowledge gained from the primary paper's observations on the decline in calibration accuracy in SSL approaches, we include our suggested approach into this comparative study, focusing on CIFAR-100 and STL10 datasets.

\end{document}